\documentclass[letterpaper,10pt,journal]{IEEEtran}

\IEEEoverridecommandlockouts  

\usepackage{amsmath}  
\usepackage{amssymb}  
\usepackage{microtype}      

\usepackage{graphicx}
\usepackage[caption=false,font=footnotesize]{subfig}
\usepackage{booktabs}       
\usepackage{tabularx}
\usepackage{array}
\usepackage{pifont}
\usepackage{makecell}

\usepackage{cite}
\usepackage{balance}
\usepackage{stfloats}

\usepackage[colorlinks = true,
  linkcolor = blue,
  urlcolor  = blue,
  citecolor = blue,
anchorcolor = blue]{hyperref}

\newcolumntype{L}[1]{>{\raggedright\arraybackslash}p{#1}} 
\newcolumntype{Y}{>{\raggedright\arraybackslash}X}

\graphicspath{%
  {./figures/}%
}

\title{\Large FUSE: A \underline{F}ramework for \underline{U}nified \underline{S}tate \underline{E}stimation \\
in Vehicular and Robotic SLAM Systems}

\author{Wei Wu$^{1}$, Honglin Chen$^{1}$, Wenhan Cao$^{1}$, Yao Lyu$^{1}$, Shaobing Xu$^{1}$, Kun Jiang$^{1}$, Jiangtao Li$^{3}$, Tao Zhang$^{3}$,\\ Lei Guo$^{4}$, Shengbo Eben Li$^{1,2}$%
  \thanks{This study is supported by National Science and Technology Major Project (No. 2025ZD1606200), Beijing Natural Science Foundation (L257002), and NSF China with 92582205. All correspondence should be addressed to S. E. Li with e-mail: lishbo@tsinghua.edu.cn.}%
  \thanks{$^{1}$W. Wu, H. Chen, W. Cao, Y. Lyu, S. Xu, K. Jiang and S. E. Li are with State Key Lab of Intelligent Green Vehicle and Mobility, Tsinghua University, Beijing 100084, China.}%
  \thanks{$^{2}$S. E. Li is also with School of VM and College of AI, Tsinghua University, Beijing 100084, China.}%
  \thanks{$^{3}$J. Li and T. Zhang are with SunRisingAI Ltd., Beijing 100082, China.}%
  \thanks{$^{4}$L. Guo is with China Intelligent and Connected Vehicles (Beijing) Research Institute Co., Ltd., Beijing 100176, China.}%
}

\begin{document}

\maketitle
\begin{abstract}
  Tightly coupled SLAM formulations under mixed-rate sensing often bind temporal processing, local geometric association, estimator formulation, and map-update policy into method-specific designs. Such binding makes it difficult to vary one design choice without re-engineering the rest of the state-estimation process. This paper presents FUSE, a framework for unified state estimation in vehicular and robotic SLAM systems. FUSE organizes the state-estimation interface around observation ingestion, propagation, update, and state query, and uses this interface to separate temporal processing, residual-ready local geometric association, estimator formulation, and map-update policy. A LiDAR--IMU instantiation is developed to examine the framework under mixed-rate sensing and directional degeneracy, where high-rate inertial propagation, LiDAR-triggered geometric update, residual screening, and degeneracy-aware correction operate through the same interface boundaries. On a 418~m loop-corridor sequence, the instantiation reports a 1.626~m end-to-end trajectory error, corresponding to a 7.9\% relative error reduction compared with Faster-LIO, the lowest-error baseline on this sequence. The results support FUSE as a framework for organizing state-estimation design choices and show how the evaluated instantiation regularizes updates along weakly observable directions.
\end{abstract}


\section{Introduction}
\label{sec:introduction}

SLAM in vehicular and robotic systems is an online state-estimation problem in which platform motion, sensor observations, and map-related variables must remain consistent under uncertainty~\cite{durrant-whyte_simultaneous_2006,cadena_past_2016}. In tightly coupled SLAM, this consistency is maintained by propagating a state using proprioceptive information and updating it with exteroceptive geometric observations~\cite{forster_imu_2015,zhang_loam_2014,shan_lio-sam_2020}. In practical vehicular and robotic systems, SLAM accuracy and consistency are challenged by mixed-rate sensing, latency-sensitive motion, and local geometric degeneracy, which complicate update timing, state availability, and the observability of certain state directions, respectively~\cite{huang_observability-based_2010,zhu_degeneracy-aware_2025}. A SLAM formulation must therefore address temporal processing, local geometric association, estimator formulation, and map-update policy as coupled state-estimation issues rather than as isolated implementation details.

Existing tightly coupled SLAM formulations often bind these issues inside method-specific designs. Temporal processing fixes how propagation, update, and query are scheduled; local geometric association determines which constraints are admitted by the estimator; estimator formulation defines the maintained state and posterior update rule; and map-update policy controls whether accepted observations modify persistent map state. Such binding can be effective for a specialized odometry formulation, but it makes estimation behavior difficult to preserve when sensing rates, spatial indices, estimator formulations, or mapping and localization requirements change~\cite{zhang_loam_2014,xu_fast-lio2_2022,he_pointlio_2023}.

This paper presents FUSE, a unified state-estimation framework for vehicular and robotic SLAM systems that organizes the problem into four interface roles (see Table~\ref{tab:fuse_coupling_summary}). The main integration challenge goes beyond decomposing temporal, spatial, inference, and functional modules into separate software blocks. It lies in preserving consistency across a temporal state history during propagation, update, and query; generating residual-ready constraints through local association without tying the estimator to a particular spatial index; encapsulating estimator-specific posterior updates behind a common state-estimation interface; and expressing mapping or localization behavior through the persistence of map updates rather than changes to the estimator itself. FUSE addresses these requirements through shared interface operations for observation ingestion, propagation, update, and query, together with a map-update policy that decouples map modification from state correction.

A LiDAR--IMU instantiation is developed to examine one concrete realization of the framework under mixed-rate sensing and directional degeneracy. In this instantiation, high-rate IMU propagation maintains queryable state history, LiDAR observations trigger geometric updates through local association, and degeneracy-aware update, residual screening, and covariance propagation are treated as realization-level mechanisms rather than as the definition of FUSE itself.

The evaluation uses a 418~m loop corridor as the primary diagnostic benchmark because its repeated corridor geometry induces weakly observable directions. On this sequence, the LiDAR--IMU instantiation reports a 1.626~m end-to-end trajectory error. Additional self-collected sequences, ablation studies, and runtime measurements are used to characterize the operating boundary of the current instantiation, not to claim categorical dominance over all LiDAR--IMU SLAM systems.

The contributions of this paper are summarized as follows:
\begin{itemize}
  \item A unified state-estimation framework that defines a common interface for observation ingestion, propagation, update, and state query.
  \item A four-role integration scheme that connects temporal history, residual-ready local association, estimator-specific updates, and map-update policy through explicit state-estimation interfaces.
  \item A LiDAR--IMU instantiation evaluated under directional degeneracy, with ablation and runtime evidence for degeneracy-aware update, residual screening, and covariance propagation.
\end{itemize}

\begin{table*}[!t]
  \centering
  \caption{State-Estimation Couplings Addressed by FUSE}
  \label{tab:fuse_coupling_summary}
  \setlength{\tabcolsep}{4pt}
  \renewcommand{\arraystretch}{1.3}
  \begin{tabularx}{\textwidth}{@{\hspace{6pt}}p{2cm}>{\raggedright\arraybackslash}p{6.8cm}>{\raggedright\arraybackslash}X@{}}
    \toprule
    \textbf{Coupling type} &
    \textbf{FUSE separation} &
    \textbf{Coupled elements} \\
    \midrule

    Temporal &
    Separates propagation, update, and query over a shared temporal state history. &
    IMU propagation--LiDAR-triggered update and query timing. \\

    Spatial &
    Exposes local associations as residual-ready constraints independent of spatial indexing. &
    Point/plane residuals--kd-tree, voxel-grid, or hash-voxel search. \\

    Inference &
    Encapsulates backend-specific posterior updates behind a common state-estimation interface. &
    Measurement residuals--KF, ESKF, IESKF, or NANO backend. \\

    Functional &
    Uses map-update policies to control map persistence without changing the estimator. &
    Accepted observations--map insertion/refinement or fixed-map query. \\

    \bottomrule
  \end{tabularx}
\end{table*}

\section{Related Work and Design Space}
\label{sec:preliminaries}

This section reviews prior work as a design space for unified state estimation rather than as a catalogue of SLAM systems. Classical SLAM surveys identify the joint estimation of platform motion and map variables as the central problem~\cite{durrant-whyte_simultaneous_2006,bailey_simultaneous_2006,cadena_past_2016}. Modern LiDAR--IMU systems achieve strong online performance by combining inertial propagation with geometric measurement updates: LOAM separates odometry and mapping through geometric features~\cite{zhang_loam_2014}, LIO-SAM formulates LiDAR--IMU estimation with smoothing and mapping factors~\cite{shan_lio-sam_2020}, FAST-LIO and FAST-LIO2 use iterated Kalman filtering with direct LiDAR updates~\cite{xu_fast-lio_2021,xu_fast-lio2_2022}, and Point-LIO and Faster-LIO further emphasize high-rate operation and efficient local association~\cite{he_pointlio_2023,bai_faster-lio_2022}. These systems are effective, but their temporal processing, local geometric association, estimator formulation, and map-update behavior are typically specified together inside a method-specific pipeline.

\begin{figure*}[!t]
  \centering
  \subfloat[Temporal processing under mixed-rate sensing.\label{fig:design_space_runtime}]{
    \includegraphics[width=0.45\textwidth,height=0.235\textheight,keepaspectratio]{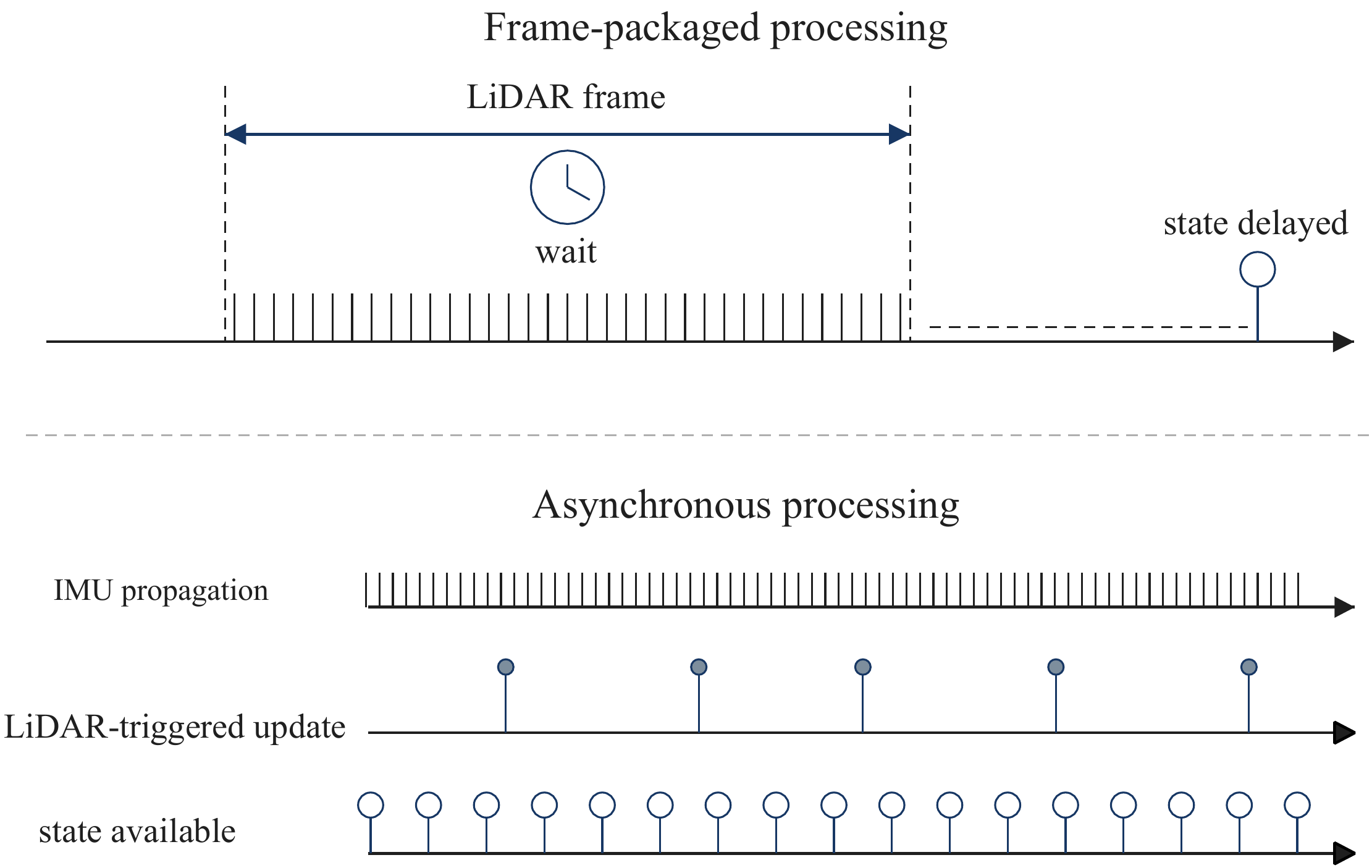}
  }
  \hspace{0.01\textwidth}
  \subfloat[Local geometric association and spatial indexing.\label{fig:design_space_spatial}]{
    \includegraphics[width=0.45\textwidth,height=0.235\textheight,keepaspectratio]{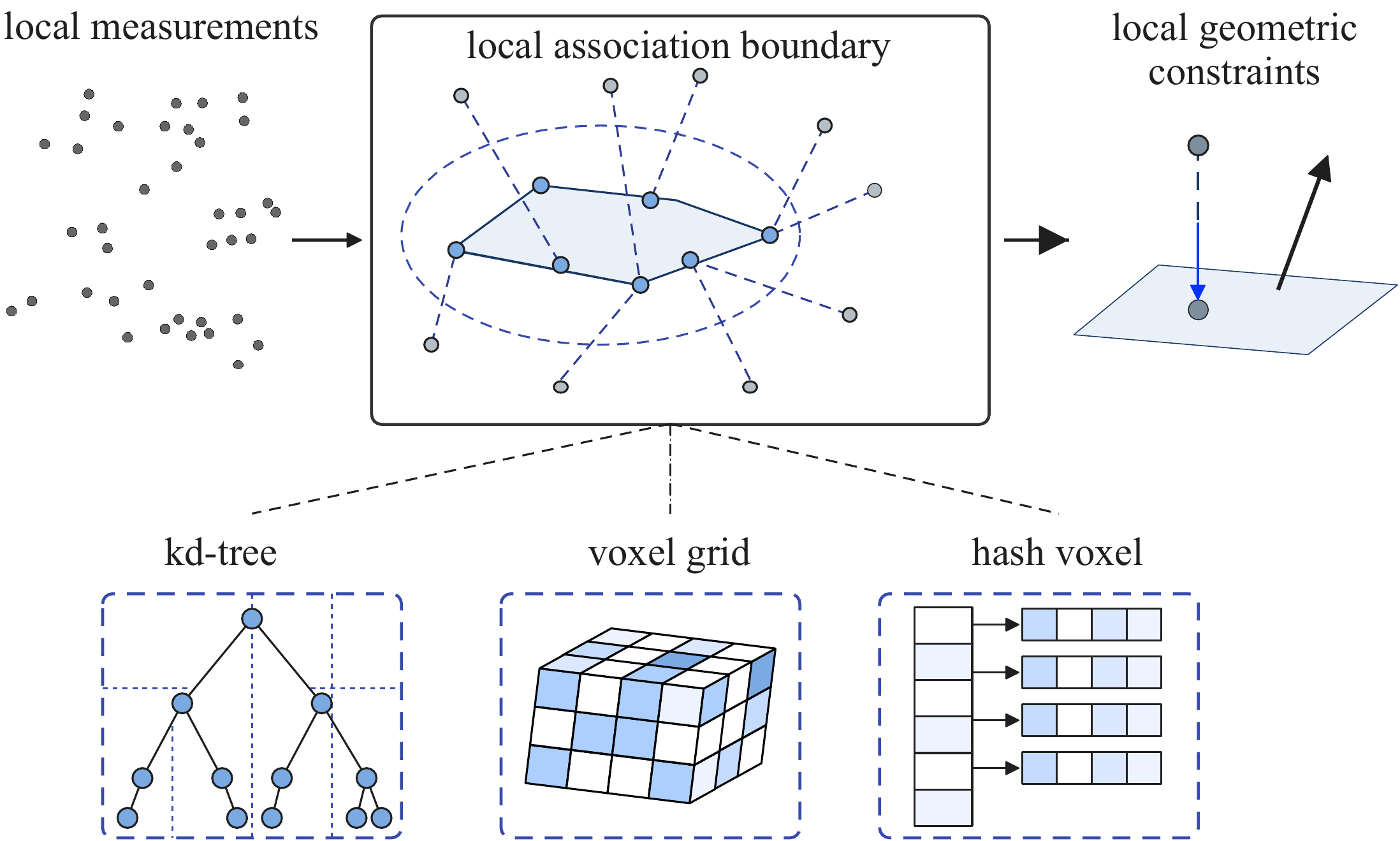}
  }

  \vspace{0.4em}

  \subfloat[Estimator formulation.\label{fig:design_space_coupling}]{
    \includegraphics[width=0.45\textwidth,height=0.235\textheight,keepaspectratio]{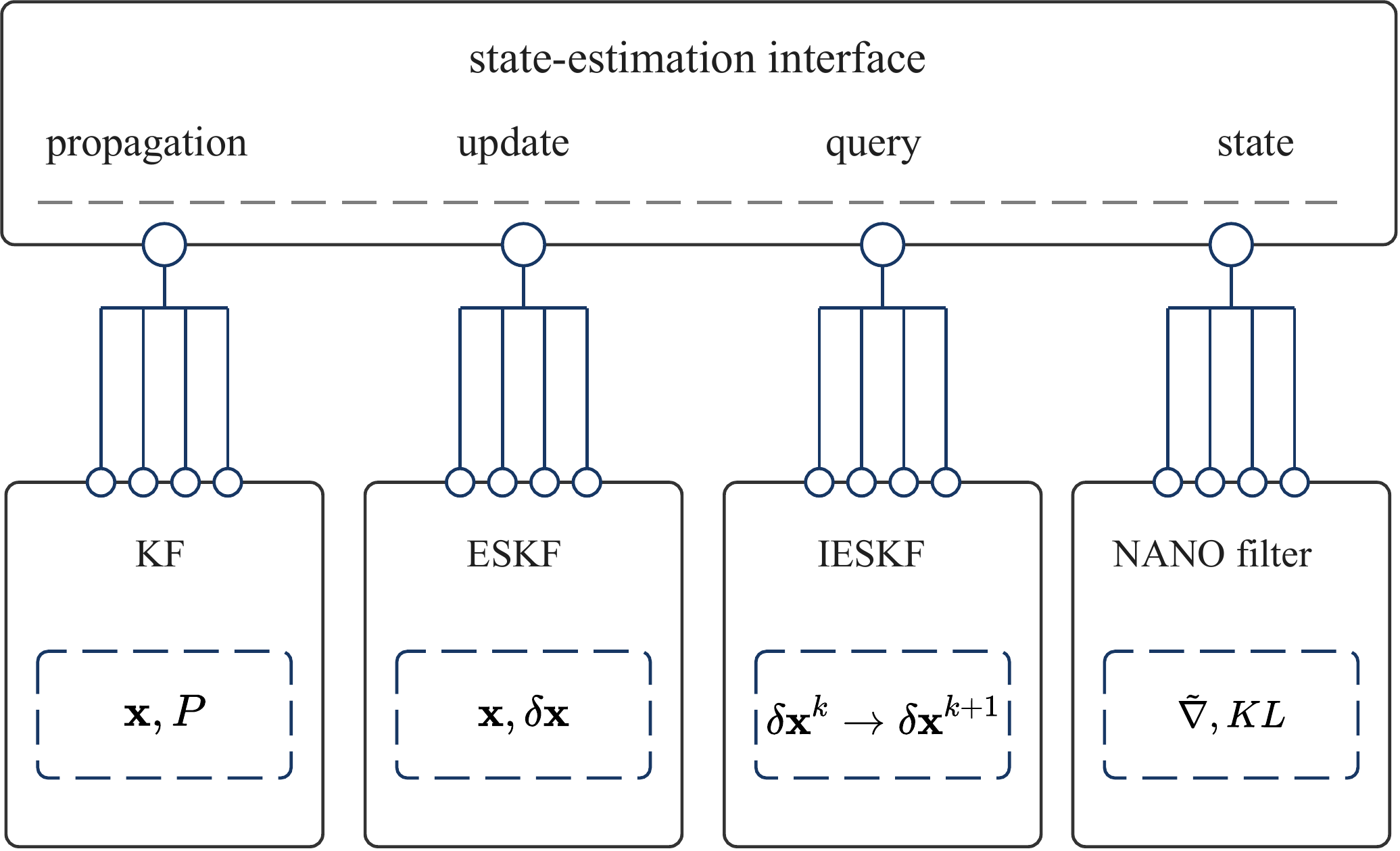}
  }
  \hspace{0.01\textwidth}
  \subfloat[Mapping and localization policies.\label{fig:design_space_map_policy}]{
    \includegraphics[width=0.45\textwidth,height=0.235\textheight,keepaspectratio]{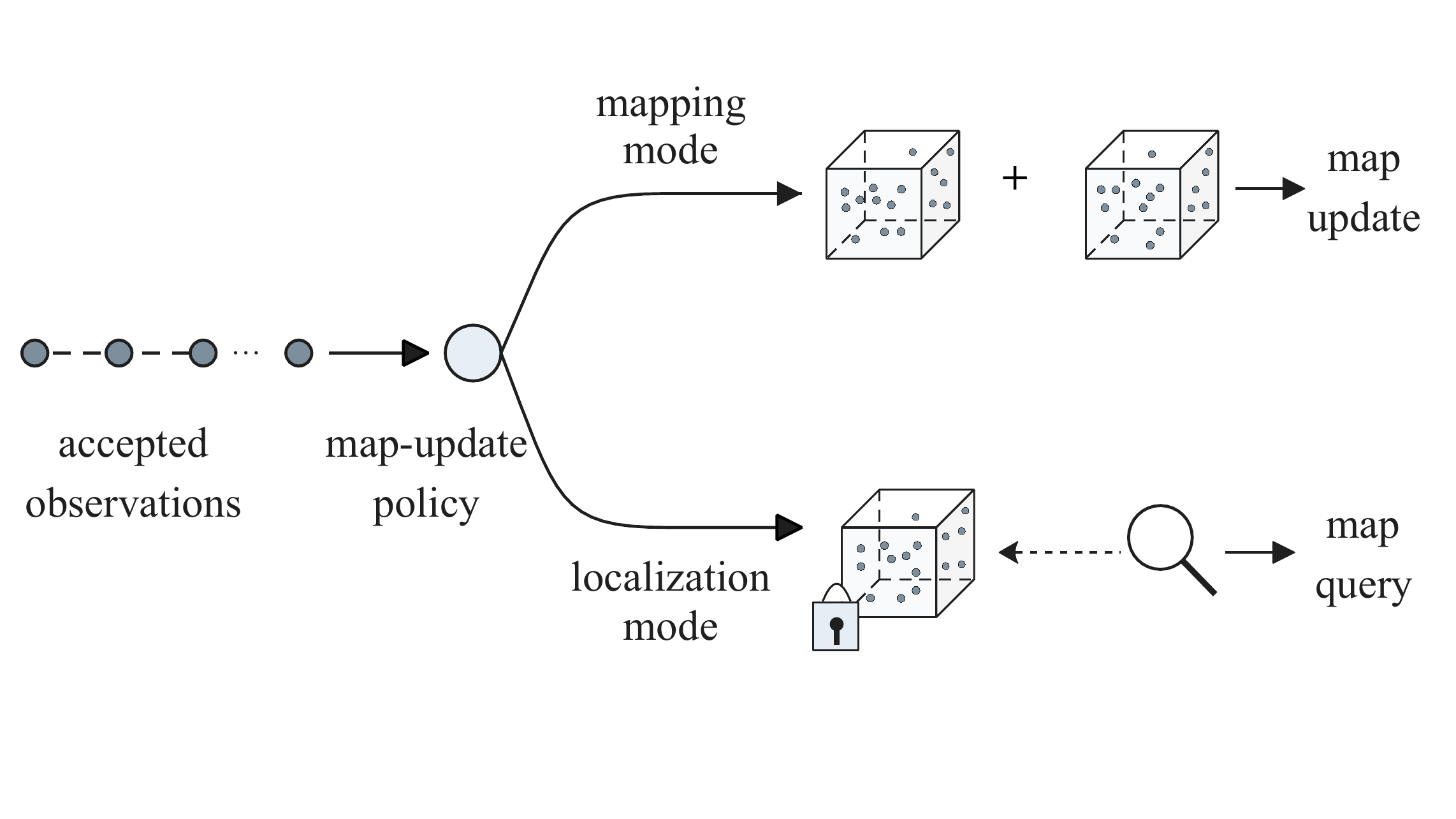}
  }
  \caption{Design space of tightly coupled LiDAR--IMU SLAM considered in this paper. The four panels correspond one-to-one to the four design aspects discussed in Section~\ref{sec:preliminaries}: temporal processing, local geometric association, estimator formulation, and map-update policy.}
  \label{fig:design_space_overview}
\end{figure*}

Multi-sensor SLAM also depends on calibration, synchronization, and source-level consistency before measurements enter the estimator~\cite{wu2023multi}. FUSE does not solve this upstream calibration problem. Instead, it assumes an ordered stream of calibrated, timestamped observations and asks how the downstream state-estimation process should expose its temporal, spatial, inference, and functional boundaries. Figure~\ref{fig:design_space_overview} summarizes these four design aspects; the subsections below explain the coupling that each aspect introduces and the interface boundary used by FUSE.

\subsection{Temporal Processing Under Mixed-Rate Sensing}

\emph{Frame-packaged processing.} A common organization packages high-rate inertial measurements around a lower-rate exteroceptive frame or scan. In visual--inertial and LiDAR--inertial systems, inertial preintegration or multi-state constraints summarize motion between camera or LiDAR frames, and the estimator performs correction after the frame is available~\cite{forster_imu_2015,mourikis_multi-state_2007,qin_vins-mono_2018}. FAST-LIO and FAST-LIO2 fit this scan-packaged organization: a LiDAR scan defines the update interval, IMU samples over that interval are buffered for propagation and deskewing, and the geometric correction is applied after the scan package has temporal support~\cite{xu_fast-lio_2021,xu_fast-lio2_2022}. This organization simplifies deskewing and data association because the update is tied to a bounded observation interval. Its limitation is that propagation, correction, and query availability tend to inherit the cadence and buffering latency of the lower-rate sensor.

\emph{Asynchronous processing.} An alternative organization does not make every correction wait for a complete exteroceptive frame. Instead, propagation, update, and query can be associated with finer-grained measurement times or continuous-time support. Point-LIO is a representative point-level LiDAR--inertial example because it updates the state at individual LiDAR point measurements rather than accumulating all points into a full frame~\cite{he_pointlio_2023}. Continuous-time spline formulations provide another way to represent states over asynchronous measurement times~\cite{mo_continuous-time_2022}. The resulting design question is not whether the sensing is tightly coupled, but whether ingestion, propagation, update, and query are exposed as separate estimation primitives. FUSE uses this separation so that temporal behavior can be specified without binding it to a particular estimator formulation, LiDAR scan package, or visual update event.

\subsection{Local Geometric Association and Spatial Indexing}

\emph{Kd-tree association.} Tree-based nearest-neighbor search is widely used to construct local geometric residuals, including point-to-plane residuals derived from neighboring map points or local surface fits~\cite{segal_generalized-icp_2009,zhang_loam_2014}. Incremental kd-tree variants reduce the cost of maintaining a changing local map and remain effective when the association procedure requires explicit nearest-neighbor neighborhoods~\cite{cai_ikd-tree_2021}. The coupling risk is that residual construction, neighborhood selection, and estimator update code can become tied to the tree representation. In the FUSE design space, a kd-tree is a possible association backend, but the estimator consumes only the residual-ready constraint set produced by that backend.

\emph{Voxel-grid association.} Voxel maps discretize space into cells that can store point summaries, local plane parameters, or probabilistic occupancy and surface statistics. Adaptive voxel mapping provides efficient online LiDAR odometry by balancing map resolution, memory footprint, and local surface representation~\cite{yuan_efficient_2022}. Voxelization can stabilize local summaries and reduce search cost, but it also introduces resolution choices and cell-level update rules that should not determine the estimator interface. FUSE therefore places voxel maintenance inside local geometric association and exposes only the resulting constraints to the update operation.

\emph{Hash-voxel association.} Hash-based voxel structures index only occupied regions and are well suited to sparse LiDAR maps and high-rate local updates. Faster-LIO uses sparse incremental voxels to reduce local association overhead, while lightweight ICP systems illustrate how simple registration choices can be competitive when the association boundary is well controlled~\cite{bai_faster-lio_2022,vizzo_kiss-icp_2023}. Hash-voxel search changes map storage and neighborhood retrieval, but it should not require a different posterior update rule. In FUSE, kd-tree, voxel-grid, and hash-voxel implementations are treated as alternative spatial indexing choices behind the same residual-generation boundary.

\subsection{Estimator Formulations}

Estimator formulations differ in the posterior representation they maintain and in the update rule used after residuals are formed. Smoothing and factor-graph methods model SLAM as nonlinear least squares over a set of states and factors, often improving delayed-state consistency at the cost of a different computational structure~\cite{kaess_isam2_2012,dellaert_factor_2017}. The present FUSE instantiation focuses on recursive filtering, but the design-space argument is broader: temporal ingestion, local association, and map-update policy should not be rewritten only because the backend posterior representation changes.

\emph{KF/EKF.} Kalman filtering maintains a Gaussian belief and updates the mean and covariance through linear or locally linearized measurement models. In SLAM, EKF consistency depends on the observability properties of the chosen state and measurement parameterization~\cite{huang_observability-based_2010}. This formulation offers a clear propagation/update interface, but its linearization and covariance representation are estimator-specific choices. FUSE therefore treats the KF or EKF update as one backend realization behind the common state-estimation interface.

\emph{ESKF.} Error-state filtering separates a nominal state from a tangent-space error state, which is useful for inertial systems on manifolds. Manifold-aware filtering and Lie-theoretic treatments define how perturbations, covariance propagation, and retraction are handled for poses, velocities, biases, and calibration parameters~\cite{barrau_invariant_2017,sola_micro_2021}. This formulation changes the internal state representation and correction operation, but it does not have to change observation ingestion, local constraint generation, or map-update policy. FUSE uses the interface boundary to keep these surrounding processes independent of the ESKF parameterization.

\emph{IESKF.} Iterated ESKF formulations repeatedly relinearize nonlinear residuals around updated nominal states, which is useful for direct LiDAR updates with point-to-plane constraints. FAST-LIO and FAST-LIO2 demonstrate that this update style can support efficient tightly coupled LiDAR--IMU odometry~\cite{xu_fast-lio_2021,xu_fast-lio2_2022}. The iteration strategy, convergence criteria, and covariance update are backend-specific. FUSE exposes the same update call to the surrounding system while allowing this internal iterative correction rule to differ from a standard EKF or ESKF update.

\emph{NANO filter.} The NANO filter reconstructs Gaussian filtering from an optimization perspective, where both the prediction and update steps are regraded as solutions to two optimization problems. The extreme condition for the problem in prediction step requires matching the first two moments of the prior distribution. In the update step, natural gradient that accounts for the curvature of the parameter space is derived to minimize the update step’s objective to avoid linearization errors in Kalman filters family~\cite{cao2026nonlinear}. NANO-SLAM applies this formulation to vehicle SLAM and motivates treating variational Gaussian updates as an estimator formulation rather than as a separate SLAM pipeline~\cite{zhang2025nano}. In FUSE, NANO filter is represented as an interface-compatible backend: it may change the posterior update operator, but it does not redefine temporal processing, local geometric association, or map-update policy. The present experiments do not claim an exhaustive comparison among all estimator backends.

\subsection{Mapping and Localization Policies}

\emph{Mapping mode.} In mapping mode, accepted observations modify persistent map state by adding, refining, or replacing geometric elements. Visual--inertial mapping frameworks, spatial-perception systems, and hierarchical map representations illustrate that maps are not only estimator outputs; they are persistent data structures with their own update and maintenance rules~\cite{schneider_maplab_2018,rosinol_kimera_2021,hughes_hydra_2022,hughes_foundations_2024}. If this logic is embedded directly in the estimator, changing from a local map to a submap or scene-graph representation can force changes to propagation and correction code. FUSE instead treats persistent map modification as a map-update policy applied after accepted observations have passed the state-estimation interface.

\emph{Localization mode.} In localization mode, observations are used primarily to query or align against an existing map, and persistent map modification is restricted or disabled. Map reuse, teach-and-repeat navigation, visual--inertial localization, and lifelong mapping all emphasize that map stability and map adaptation are operational choices rather than consequences of a different estimator definition~\cite{sun_robust_2021,campos_orb-slam3_2021,banerjee_lifelong_2023}. FUSE expresses this distinction by changing the map-update policy while preserving the same propagation, local association, update, and query operations. Mapping and localization are therefore represented as different persistence policies, not as separate SLAM pipelines.

Directional degeneracy is a realization-level stressor rather than a fifth interface role. Degeneracy-aware LiDAR--inertial odometry and constraint-based mapping studies show that weakly observable environments require safeguards on the admitted residuals and on the update directions~\cite{lim_adalio_2023,zhu_degeneracy-aware_2025,wu2025enhancing}. In FUSE, such safeguards are attached to the local association and estimator-update realization, while the framework-level interface remains the same. This distinction motivates the LiDAR--IMU instantiation in Section~\ref{sec:lio_instantiation}, where residual screening and directional update regularization are evaluated under corridor degeneracy.

Collectively, the four design aspects in Fig.~\ref{fig:design_space_overview} identify where tightly coupled SLAM systems often bind choices that could be exposed through interfaces. FUSE does not claim that every temporal schedule, spatial index, estimator backend, or map policy has been experimentally validated. Its claim is narrower: these choices can be organized around a shared state-estimation interface so that a concrete LiDAR--IMU instantiation can vary realization-level mechanisms without redefining the overall state-estimation process.

\section{FUSE: A Unified State-Estimation Framework}
\label{sec:fuse_framework}

\begin{figure*}[!t]
  \centering
  \includegraphics[width=0.95\textwidth]{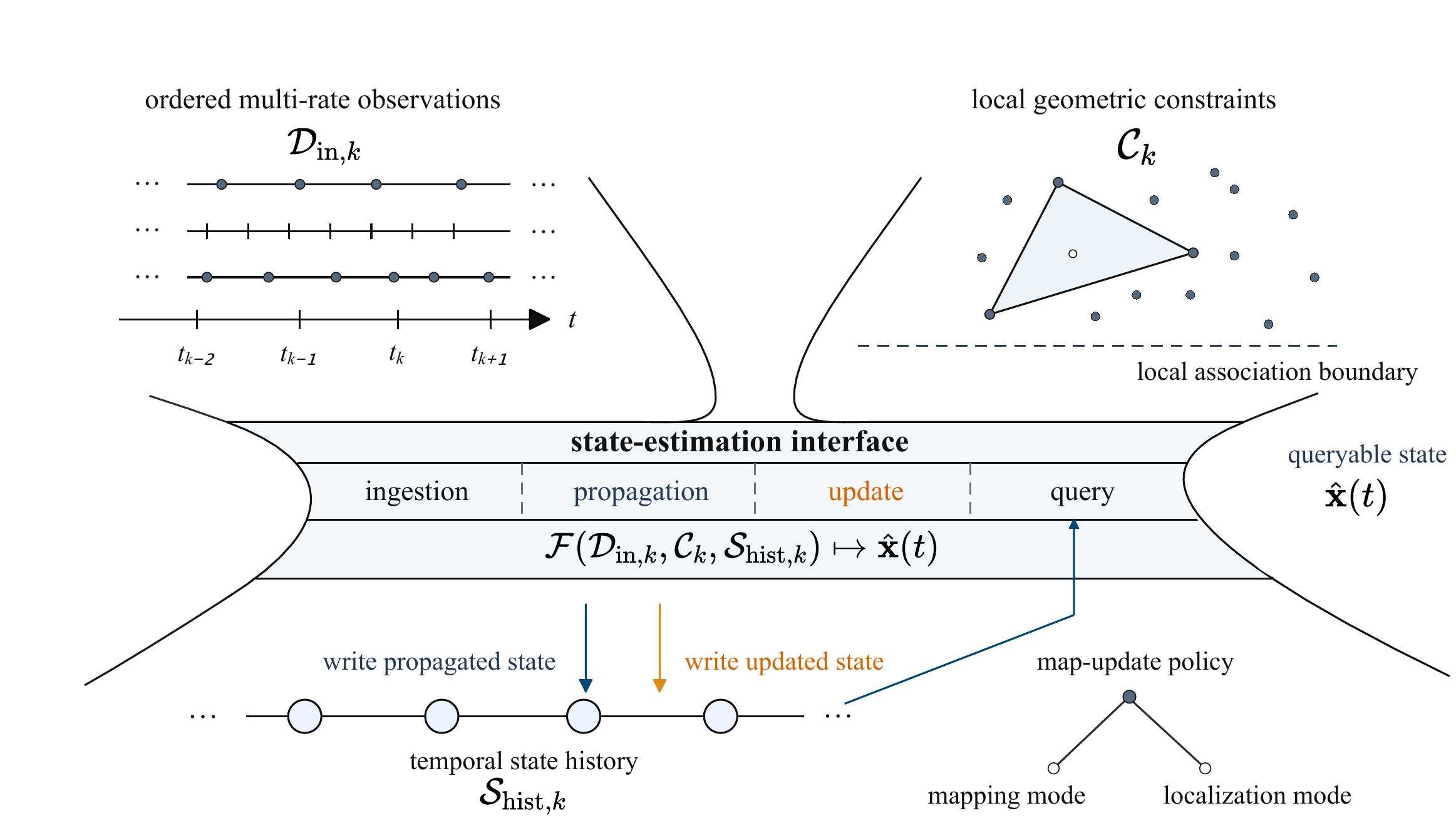}
  \caption{Conceptual framework of FUSE for unified state estimation in vehicular and robotic SLAM systems. Ordered multi-rate observations and residual-ready local geometric constraints enter the state-estimation interface, which exposes ingestion, propagation, update, and query operations. Propagation and update write propagated or corrected beliefs into the temporal state history, while query returns a state estimate $\hat{\mathbf{x}}(t)$ at a requested time. Estimator formulations implement posterior updates behind the interface, and the map-update policy determines whether accepted observations modify persistent map state or only support fixed-map query.}
  \label{fig:fuse_framework}
\end{figure*}

FUSE specifies a state-estimation framework that separates four interface roles: temporal processing, local geometric association, estimator formulation, and map-update policy. The framework is organized around observation ingestion, propagation, update, and query operations. These operations define how timestamped observations enter the estimator, how propagated and updated beliefs are written into temporal state history, how residual-ready local constraints are consumed by an update backend, and how accepted observations affect persistent map state. Figure~\ref{fig:fuse_framework} summarizes this organization. The purpose of the framework is not to make all design choices interchangeable without implementation effort, but to make their interaction points explicit so that temporal scheduling, spatial indexing, posterior update rules, and map persistence are not hard-coded into a single SLAM pipeline.

\subsection{Problem Formulation}

We define an ordered, heterogeneous observation stream as a sequence of timestamped measurements:
\begin{equation}
  \mathcal{D}_{\mathrm{in},k}
  =
  \left\{
    \left(\mathbf{z}_i,t_i,s_i\right)
  \right\}_{i=1}^{k},
  \qquad
  t_1 < t_2 < \cdots < t_k ,
  \label{eq:fuse_ordered_input}
\end{equation}
where each tuple comprises a measurement vector $\mathbf{z}_i \in \mathbb{R}^n$, a timestamp $t_i$, and a source identifier $s_i$. The stream may interleave high-rate proprioceptive measurements with lower-rate exteroceptive observations. The estimation objective is to maintain a posterior over the platform state and map-related variables while keeping the state queryable at the times required by downstream operations. FUSE treats this objective as an interface problem: the estimator should expose the operations needed for propagation, update, and query without binding them to a fixed sensor schedule, a specific spatial index, or one posterior-update formulation.

To support mixed-rate sensing, FUSE maintains a temporal state history:
\begin{equation}
  \mathcal{S}_{\mathrm{hist},k}
  =
  \left\{
    \left(\tau_j,\hat{\mathbf{x}}_j,\mathcal{B}_j\right)
    \mid
    \tau_j \leq t_k
  \right\},
\end{equation}
where $\hat{\mathbf{x}}_j$ is a manifold-valued state estimate at time $\tau_j$, and $\mathcal{B}_j$ denotes the associated stochastic belief, such as a covariance or information matrix. The history provides the temporal support for propagated-state queries, delayed observation handling, and motion compensation of exteroceptive measurements. It also defines the shared time base on which propagation, update, and query operations must remain consistent.

\subsection{State-Estimation Interface}

The state-estimation interface defines the common inputs and outputs used by the four framework roles. It receives the ordered observation stream, local geometric constraints generated by association, and the temporal state history. The interface is summarized by the mapping
\begin{equation}
  \mathcal{F}:
  \left(
    \mathcal{D}_{\mathrm{in},k},
    \mathcal{C}_{k},
    \mathcal{S}_{\mathrm{hist},k}
  \right)
  \mapsto
  \hat{\mathbf{x}}(t),
  \label{eq:fuse_estimator_boundary}
\end{equation}
where $\mathcal{C}_{k}$ denotes the set of geometric correspondences or constraints yielded by local association during update cycles.

The mapping in \eqref{eq:fuse_estimator_boundary} is a boundary specification rather than a particular estimator. It does not prescribe state dimension, residual parameterization, spatial index, or uncertainty representation. These choices are assigned to the corresponding realization: temporal processing determines when propagation and update occur; local geometric association determines how $\mathcal{C}_{k}$ is produced; the estimator formulation determines how posterior updates are computed; and the map-update policy determines whether accepted observations change persistent map state. This separation makes the coupling points explicit without claiming that every backend or map representation has been exhaustively validated in this paper.

\subsection{Asynchronous Propagation and Update}

Temporal processing separates state evolution from measurement-driven correction. The propagation operator $\Phi_k$ advances the state belief using motion-related measurements:
\begin{equation}
  \left(
    \hat{\mathbf{x}}_{k+1}^{-},
    \mathcal{B}_{k+1}^{-}
  \right)
  =
  \Phi_k
  \left(
    \hat{\mathbf{x}}_{k}^{+},
    \mathcal{B}_{k}^{+},
    \mathcal{U}_{k:k+1}
  \right),
  \label{eq:fuse_prediction}
\end{equation}
where $\mathcal{U}_{k:k+1}$ denotes the proprioceptive sequence integrated between update instants. The superscripts ${}^-$ and ${}^+$ denote propagated and updated beliefs, respectively.

The update operator $\Gamma_k$ incorporates exteroceptive evidence once the required temporal support and geometric constraints are available:
\begin{equation}
  \left(
    \hat{\mathbf{x}}_{k}^{+},
    \mathcal{B}_{k}^{+}
  \right)
  =
  \Gamma_k
  \left(
    \hat{\mathbf{x}}_{k}^{-},
    \mathcal{B}_{k}^{-},
    \mathcal{Z}_{k},
    \mathcal{C}_{k},
    \mathcal{S}_{\mathrm{hist},k}
  \right),
  \label{eq:fuse_correction}
\end{equation}
where $\mathcal{Z}_{k}$ is the set of observations eligible for innovation. Equations~\eqref{eq:fuse_prediction} and \eqref{eq:fuse_correction} express the scheduling boundary used by FUSE. For example, in a LiDAR--IMU instantiation, IMU data may drive high-rate propagation while LiDAR frames trigger lower-rate geometric updates; the same boundary can be retained even when the backend update rule changes.

\subsection{Local Geometric Association}

Local geometric association converts exteroceptive measurements and local map information into constraints that can be consumed by the estimator. FUSE represents this role by the operator $\mathcal{A}$:
\begin{equation}
  \mathcal{A}
  \left(
    \mathcal{M}_{\mathrm{loc},k},
    \mathbf{z}_{k},
    \hat{\mathbf{x}}_{k}^{-}
  \right)
  \mapsto
  \mathcal{C}_{k},
  \label{eq:fuse_association}
\end{equation}
where $\mathcal{M}_{\mathrm{loc},k}$ is the local map representation, $\mathbf{z}_{k}$ is the incoming geometric observation, and $\hat{\mathbf{x}}_{k}^{-}$ is the propagated state used to seed association.

The output $\mathcal{C}_{k} = \left \{ c_{j,k} \right \}_{j=1}^{m_k}$ is a set of residual-ready correspondences or geometric constraints. Each element $c_{j,k}$ defines a residual $r_{j,k}(\mathbf{x}) = r\left(c_{j,k},\mathbf{x}\right)$ exposed to the update operator. The estimator therefore consumes the constraint set, not the spatial index that produced it. A kd-tree, voxel grid, hash-voxel map, or plane-summary structure may be used inside association, but the posterior update is expressed through $\mathcal{C}_{k}$.

\subsection{Map-Update Policy}

Map management is represented as a map-update policy rather than as a separate estimator pipeline. Let $\mathcal{M}_{k}$ denote the persistent map or submap state, and let $\mathcal{K}_{k}$ denote the accepted constraints or observations after estimator update and admissibility checks. The map state evolves according to a mode-dependent policy:
\begin{equation}
  \mathcal{M}_{k+1}
  =
  \begin{cases}
    \Psi\left(\mathcal{M}_{k}, \mathcal{K}_{k}\right), & \text{mapping mode}, \\
    \Omega\left(\mathcal{M}_{k}, \mathcal{K}_{k}\right), & \text{localization mode},
  \end{cases}
  \label{eq:fuse_map_policy}
\end{equation}
where $\Psi$ represents map insertion, replacement, or refinement, and $\Omega$ represents fixed-map query or restricted modification.

Equation~\eqref{eq:fuse_map_policy} separates map persistence from state correction. Mapping and localization can therefore share the same propagation, association, update, and query operations while differing in whether accepted observations modify the persistent map. This distinction is important for keeping functional mode changes from altering the estimator formulation.

\subsection{Estimator Formulations}

Estimator formulations define the posterior representation and the update law that operate behind $\Gamma_k$ in \eqref{eq:fuse_correction}. FUSE treats KF, ESKF, IESKF, and NANO-style updates as possible backend formulations under the same ingestion, propagation, update, and query boundary.

For Gaussian-based formulations, the maintained belief can be written as
\begin{equation}
  q_k(\mathbf{x})
  =
  \mathcal{N}
  \left(
    \mathbf{x};
    \hat{\mathbf{x}}_k,
    \mathbf{P}_k
  \right),
\end{equation}
where the state $\mathbf{x}$ may be represented in a Euclidean vector space or on a differentiable manifold. A standard KF assumes linear-Gaussian dynamics and observations, whereas ESKF and IESKF formulations represent corrections in a tangent space around a nominal state; IESKF additionally uses iterative relinearization for nonlinear residuals. These differences affect the internal update rule but not the surrounding interface roles.

The NANO filter introduces a variational Gaussian update on a statistical manifold through KL-divergence minimization and natural gradients. In the FUSE framework, this changes the internal realization of $\Gamma_k$ in \eqref{eq:fuse_correction}. Observation ingestion, temporal propagation, local geometric association, and map-update policy remain defined by the same interface boundary. The current experiments evaluate one LiDAR--IMU realization and do not constitute an exhaustive comparison across all estimator formulations.

FUSE is therefore a framework for organizing state-estimation interfaces rather than a single estimator or a specific LiDAR--IMU odometry algorithm. Section~\ref{sec:lio_instantiation} gives one LiDAR--IMU instantiation of the framework. In that instantiation, asynchronous IMU propagation, LiDAR-triggered local association, degeneracy-aware update, residual screening, covariance propagation, and map-update policy are realization-level mechanisms used to examine the framework under mixed-rate sensing and directional degeneracy.

\section{LiDAR--IMU Instantiation}
\label{sec:lio_instantiation}

This section gives one LiDAR--IMU instantiation of the FUSE framework defined in Section~\ref{sec:fuse_framework}. The instantiation maps the four interface roles to concrete operations: high-rate IMU propagation and LiDAR-triggered correction realize temporal processing; local map association produces residual-ready geometric constraints; the tangent-space update implements one estimator backend behind the state-estimation interface; and the map-update policy separates mapping from localization behavior. Degeneracy-aware update, residual screening, and covariance propagation are used here as realization-level safeguards for corridor-like directional degeneracy. They are not additional framework layers and do not define FUSE itself.

\begin{figure*}[!t]
  \centering
  \includegraphics[width=0.95\textwidth]{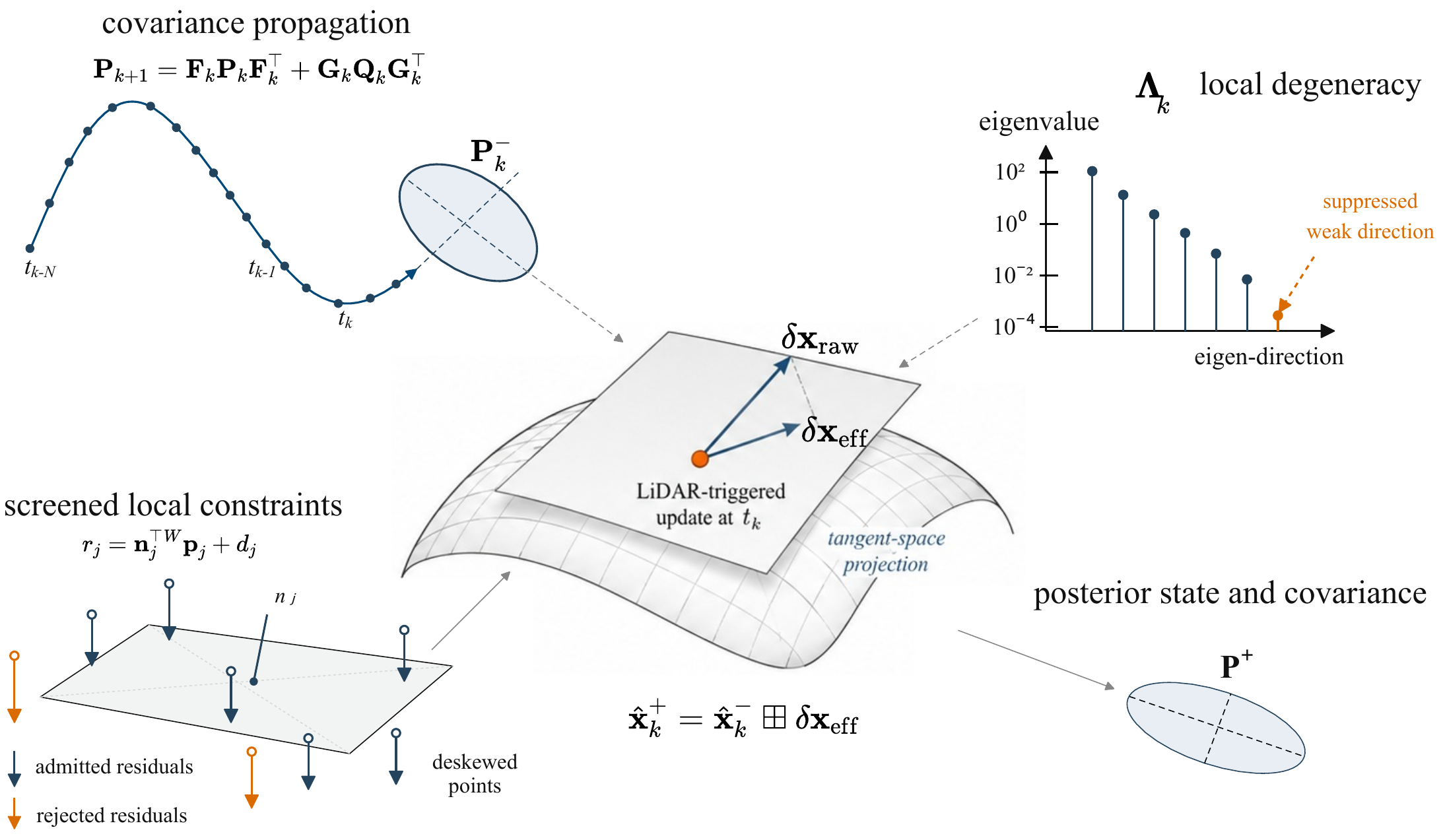}
  \caption{Internal logic of the FUSE LiDAR--IMU instantiation under mixed-rate sensing and directional degeneracy. IMU measurements propagate the manifold state and covariance, while temporal state history supports query-based LiDAR deskewing. Local association converts deskewed points into screened point-to-plane residuals. The admitted residuals form a local information matrix used to attenuate tangent-space corrections along rank-deficient directions, and the map-update policy determines whether the accepted observations update a persistent map or support fixed-map localization.}
  \label{fig:fuse_lio_instantiation}
\end{figure*}

Figure~\ref{fig:fuse_lio_instantiation} summarizes the estimation flow. IMU measurements propagate the manifold state and maintain temporal history. LiDAR observations are deskewed by querying this history, associated with local map primitives, filtered by admissibility tests, and then used in a LiDAR-triggered tangent-space update. The information matrix of the admitted constraints is used to attenuate correction components along weakly observable directions, while the map-update policy determines whether accepted observations modify the persistent map.

\subsection{State and Measurement Models}

The LiDAR--IMU estimator is defined over the world ($W$), IMU ($I$), and LiDAR ($L$) frames. We define the system state on a compound manifold $\mathcal{X}$ as the tuple:
\begin{equation}
  \mathbf{x} =
  \left(
    {}^{W}\mathbf{p}_{I},
    {}^{W}\mathbf{R}_{I},
    {}^{I}\mathbf{p}_{L},
    {}^{I}\mathbf{R}_{L},
    {}^{W}\mathbf{v}_{I},
    \mathbf{b}_{g},
    \mathbf{b}_{a},
    \mathbf{g}
  \right) \in \mathcal{X},
\end{equation}
where ${}^{W}\mathbf{p}_{I}$ and ${}^{W}\mathbf{R}_{I}$ are the IMU position and orientation in the world frame, ${}^{I}\mathbf{p}_{L}$ and ${}^{I}\mathbf{R}_{L}$ are the LiDAR--IMU extrinsic parameters, ${}^{W}\mathbf{v}_{I}$ is the IMU velocity, $\mathbf{b}_{g}$ and $\mathbf{b}_{a}$ are gyroscope and accelerometer biases, and $\mathbf{g}\in\mathbb{S}^{2}$ denotes the gravity direction. The update is applied through a tangent perturbation $\delta\mathbf{x}$ on $\mathcal{X}$ using the manifold operation $\boxplus$.

For a deskewed LiDAR point ${}^{L}\mathbf{p}$, the predicted point position in the world frame is
\begin{equation}
  {}^{W}\mathbf{p}
  =
  {}^{W}\mathbf{R}_{I}
  \left(
    {}^{I}\mathbf{R}_{L}\,{}^{L}\mathbf{p}
    +
    {}^{I}\mathbf{p}_{L}
  \right)
  +
  {}^{W}\mathbf{p}_{I}.
\end{equation}
The LiDAR measurement model uses point-to-plane constraints derived from local map associations. For a planar surface characterized by a unit normal $\mathbf{n}$ and distance $d$, the observation residual is
\begin{equation}
  r(\mathbf{x};{}^{L}\mathbf{p})=\mathbf{n}^{\top}{}^{W}\mathbf{p}+d.
\end{equation}
Applying a first-order Taylor expansion around the nominal state $\hat{\mathbf{x}}$ yields
\begin{equation}
  r(\hat{\mathbf{x}}\boxplus \delta \mathbf{x})
  \approx
  r(\hat{\mathbf{x}})+\mathbf{H}\delta \mathbf{x},
\end{equation}
where $\mathbf{H}$ denotes the measurement Jacobian with respect to the tangent perturbation. This point-to-plane model is used as a conventional LiDAR geometric constraint for the present instantiation. Within FUSE, its role is to generate residual-ready constraints for the update interface rather than to define the framework or restrict it to one association model.

\subsection{IMU Propagation and LiDAR Update}

High-rate IMU measurements propagate the nominal state and covariance. The resulting estimates are written into a temporal state history, $\mathcal{S}_{\mathrm{hist},k}$, which provides queryable states for LiDAR deskewing and for downstream state requests. This operation instantiates the propagation operator in \eqref{eq:fuse_prediction} for the LiDAR--IMU manifold state.

When a LiDAR scan arrives, the update is delayed only until the state history spans the timestamps required for motion compensation. Each point is deskewed by querying $\mathcal{S}_{\mathrm{hist},k}$, projected into the world frame, and associated with local map primitives. The accepted associations form the constraint set used by the LiDAR-triggered update in \eqref{eq:fuse_correction}. The LiDAR frame rate therefore determines the correction rate, while IMU propagation keeps the state queryable between LiDAR updates.

This interface separation prevents the state-query path from being tied to the LiDAR update cycle. Propagated estimates can be returned before the next LiDAR correction, and geometric information is incorporated once deskewing, association, and screening have produced admissible constraints. This is the scheduling behavior evaluated by the current instantiation; hard real-time guarantees are limited to the runtime profile reported in Section~\ref{sec:experiments}.

\subsection{Local Map Association}

Local map association transforms deskewed LiDAR observations and the propagated state into a structured set of geometric constraints. To instantiate the abstract operator $\mathcal{A}(\cdot)$ in \eqref{eq:fuse_association}, we define the concrete LiDAR constraint set as
\begin{equation}
  \mathcal{C}_{k}
  =
  \left\{
    \left({}^{L}\mathbf{p}_{j,k},\mathbf{n}_{j,k},d_{j,k},q_{j,k}\right)
  \right\}_{j=1}^{m_k},
\end{equation}
where each tuple comprises a motion-compensated point ${}^{L}\mathbf{p}_{j,k}$, the normal $\mathbf{n}_{j,k}$ and offset $d_{j,k}$ of the associated planar primitive, and a quality metric $q_{j,k}$ used for residual screening.

The spatial index is an implementation choice inside this association step. Tree-based nearest-neighbor search, voxel-centered summaries, or hash-based structures may be used to retrieve local primitives, but the estimator receives the same type of constraint set $\mathcal{C}_{k}$. The instantiation therefore keeps index-specific search and maintenance outside the estimator backend, while allowing the update to consume residuals and Jacobians derived from the accepted constraints.

\subsection{Degeneracy-Aware Update}

Directional degeneracy arises when the local geometric constraints provide limited information along one or more tangent-space directions. In corridor-like or locally planar environments, point-to-plane residuals can strongly constrain motion transverse to dominant surfaces while leaving motion along the principal direction weakly determined. The update therefore examines the spectral structure of the admitted constraints before applying the tangent-space correction.

The collective information represented by the admitted constraint set $\mathcal{I}_{k}$ is approximated by the normal matrix
\begin{equation}
  \boldsymbol{\Lambda}_{k}=\sum_{j\in \mathcal{I}_{k}} w_{j}\mathbf{H}_{j}^{\top}\mathbf{H}_{j},
\end{equation}
where the weights $w_j$ incorporate residual confidence. The matrix $\boldsymbol{\Lambda}_{k}$ provides a local Fisher-information approximation for the LiDAR constraints in the tangent space.

To identify weakly constrained directions, we use the eigendecomposition
\begin{equation}
  \boldsymbol{\Lambda}_{k}=\mathbf{V}\mathbf{D}\mathbf{V}^{\top}.
\end{equation}
Eigenvectors associated with small eigenvalues indicate local rank deficiency. The raw update $\delta \mathbf{x}^{\star}$ is then projected through a directional gating operator $\boldsymbol{\Gamma}$:
\begin{equation}
  \delta \mathbf{x}_{\mathrm{eff}}
  =
  \boldsymbol{\Gamma}\,\delta \mathbf{x}^{\star},
  \qquad
  \boldsymbol{\Gamma}=\mathbf{V}\,\mathrm{diag}(\gamma_{1},\ldots,\gamma_{n})\,\mathbf{V}^{\top},
\end{equation}
where the attenuation coefficients $\gamma_{i}\in[0,1]$ suppress correction components along ill-conditioned directions and preserve components supported by sufficient local information.

This mechanism is a safeguard on the update, not a recovery of missing information. It does not remove the environmental degeneracy; instead, it limits the influence of rank-deficient constraints on the state increment. The ablation study later evaluates this mechanism as part of the LiDAR--IMU instantiation rather than as a framework-level requirement.

\subsection{Propagation Covariance and Residual Screening}

Propagation covariance describes the uncertainty of the IMU-driven predictive state before the LiDAR-triggered update. We characterize the discrete-time tangent-space error dynamics as
\begin{equation}
  \delta \mathbf{x}_{k+1}=\mathbf{F}_{k}\delta \mathbf{x}_{k}+\mathbf{G}_{k}\mathbf{w}_{k},
\end{equation}
with covariance propagation
\begin{equation}
  \mathbf{P}_{k+1}
  =
  \mathbf{F}_{k}\mathbf{P}_{k}\mathbf{F}_{k}^{\top}
  +
  \mathbf{G}_{k}\mathbf{Q}_{k}\mathbf{G}_{k}^{\top},
\end{equation}
where $\mathbf{Q}_{k}$ is the discrete process noise integrated over the propagation interval. The propagated covariance supplies the prior uncertainty used by the subsequent update and controls the relative weighting between inertial prediction and LiDAR residuals.

Before correction, residual screening filters candidate correspondences returned by local map association. Let $\mathcal{I}^{\mathrm{raw}}_{k}$ denote the raw constraint indices. The admitted subset is
\begin{equation}
  \mathcal{I}_{k}
  =
  \left\{
    j\in \mathcal{I}^{\mathrm{raw}}_{k}
    \;\middle|\;
    q_{j}\ge \tau_{q},\;
    |r_{j}|\le \tau_{r}
  \right\},
\end{equation}
where $q_{j}$ is a structural quality metric and $r_{j}$ is the geometric residual. Constraints that fail these admissibility checks are not used in the LiDAR update.

Covariance propagation and residual screening act at different points of the realization. The former represents prior uncertainty accumulated during IMU propagation, while the latter controls which geometric constraints are allowed to form the innovation. Both mechanisms operate behind the same propagation and update interfaces and therefore do not alter the framework definition.

\subsection{NANO Backend Formulation}

The NANO filter can be represented as an interface-compatible estimator formulation within FUSE. It treats nonlinear Bayesian filtering as Gaussian approximation on a statistical manifold, with natural-gradient updates derived from variational inference~\cite{cao2026nonlinear}. In the LiDAR--IMU tangent space centered at the propagated nominal state $\hat{\mathbf{x}}_{k}^{-}$, the local posterior approximation is
\begin{equation}
  q_{k}(\delta\mathbf{x})
  =
  \mathcal{N}
  \left(
    \delta\mathbf{x};
    \boldsymbol{\mu}_{k},
    \mathbf{P}_{k}
  \right).
\end{equation}
Starting from the propagated Gaussian prior $q_{k}^{-}(\delta\mathbf{x}) = \mathcal{N} \left(\delta\mathbf{x}; \mathbf{0}, \mathbf{P}_{k}^{-} \right)$, and given the LiDAR likelihood induced by the admitted constraints $\mathcal{C}_k$, the optimal Gaussian posterior $q^{\star}_k$ is sought by minimizing the energy functional $J_k(\boldsymbol{\mu}_{k},\mathbf{P}_{k})$, which balances the prior divergence against the expected log-likelihood:
\begin{equation}
  \begin{aligned}
    J_{k}(\boldsymbol{\mu}_{k},\mathbf{P}_{k})
    &=
    D_{\mathrm{KL}}
    \left(
      \mathcal{N}(\boldsymbol{\mu}_{k},\mathbf{P}_{k})
      \,\middle\|\,
      \mathcal{N}(\mathbf{0},\mathbf{P}_{k}^{-})
    \right) \\
    &\quad
    -
    \mathbb{E}_{\mathcal{N}(\boldsymbol{\mu}_{k},\mathbf{P}_{k})}
    \left[
      \log p
      \left(
        \mathcal{Z}_{k}
        \mid
        \hat{\mathbf{x}}_{k}^{-}\boxplus\delta\mathbf{x},
        \mathcal{C}_{k}
      \right)
    \right].
  \end{aligned}
\end{equation}

The NANO update iteratively refines the Gaussian parameters using natural-gradient descent:
\begin{equation}
  \Delta\boldsymbol{\eta}_{k}
  =
  -
  \mathbf{F}_{\boldsymbol{\eta}}^{-1}
  \nabla_{\boldsymbol{\eta}} J_{k},
\end{equation}
where $\boldsymbol{\eta}_{k}$ denotes the mean and  covariance of the Gaussian distribution and $\mathbf{F}_{\boldsymbol{\eta}}$ is the corresponding Fisher Information Matrix. Upon convergence, the manifold state and covariance are updated as $\hat{\mathbf{x}}_{k}^{+} = \hat{\mathbf{x}}_{k}^{-}\boxplus\boldsymbol{\mu}_{k}^{\star}$ and $\mathbf{P}_{k}^{+} = \mathbf{P}_{k}^{\star}$.

Under the FUSE interface, this formulation changes the internal update operator but not observation ingestion, temporal propagation, local association, or map-update policy. The NANO-SLAM formulation~\cite{zhang2025nano} is therefore discussed as an estimator formulation compatible with the same update boundary. The experiments in this paper focus on the evaluated LiDAR--IMU realization and do not constitute an exhaustive comparison among estimator backends.

\subsection{Mapping and Localization Policy}

The LiDAR--IMU instantiation uses the map-update policy in \eqref{eq:fuse_map_policy} to distinguish mapping from localization. Both modes use the same propagation, local association, residual screening, and update interfaces. The difference is whether the accepted observation set $\mathcal{K}_{k}$ is allowed to modify persistent map state.

In mapping mode, accepted observations expand or refine the map after the estimator update. In localization mode, the system queries an existing map to support state correction while inhibiting persistent map modification. This policy-level distinction avoids rewriting the estimator when switching between map construction and map reuse.

The mechanisms in this section therefore instantiate the four framework roles without turning FUSE into a single LiDAR--IMU odometry algorithm. IMU propagation realizes the temporal role, local association provides residual-ready constraints, the estimator backend applies tangent-space posterior correction, and the map-update policy controls persistence. Section~\ref{sec:experiments} evaluates this particular realization under mixed-rate sensing and directional degeneracy.

\section{Experiments}
\label{sec:experiments}

This section evaluates the current LiDAR--IMU instantiation of FUSE. The evaluation is organized around three questions: how the instantiation behaves in a diagnostic loop-corridor sequence with directional degeneracy, what operating boundary is observed across additional self-collected sequences, and how the realization-level mechanisms in Section~\ref{sec:lio_instantiation} affect the reported corridor result. The results should be read as evidence for this instantiation under the reported sensor rates, hardware, and benchmark conditions, rather than as exhaustive validation of all temporal schedules, spatial indexes, estimator formulations, or map-update policies supported by the framework.

\begin{table*}[!b]
  \centering
  \caption{Quantitative results on the self-collected benchmark. End-to-end trajectory error is reported in meters, and lower values indicate better performance. Values are reported to three decimals to preserve the numerical output of the evaluation, not to imply millimeter-level measurement accuracy. Values within \(0.01~\mathrm{m}\) of the lowest reported error in a row are treated as centimeter-level ties.}
  \label{tab:benchmark_results}
  \small
  \setlength{\tabcolsep}{4pt}
  \renewcommand{\arraystretch}{1.0}
  \begin{tabular*}{\textwidth}{@{\extracolsep{\fill}} p{0.12\textwidth} p{0.18\textwidth} c c c c c c c}
    \toprule
    Sequence & Scene category & Time & Dist. & FAST-LIO2 & LIO-SAM & Point-LIO & Faster-LIO & FUSE \\
    &                & (min:sec) & (km) &  &  &  &  &  \\
    \midrule
    \multicolumn{9}{l}{\textit{Primary diagnostic benchmark}} \\
    \midrule
    Test1
    & loop corridor
    & 06:20 & 0.418
    & 8.189 & Fail. & 8.077 & 1.765 & \textbf{1.626} \\
    \midrule
    \multicolumn{9}{l}{\textit{Broader benchmark coverage and operating boundary}} \\
    \midrule
    Test2
    & multi-floor stairs
    & 05:34 & 0.257
    & 0.068 & Fail. & \textbf{0.007} & \textbf{0.005} & \textbf{0.005} \\
    Test3
    & campus walkway
    & 13:40 & 1.067
    & 5.468 & \textbf{4.730} & 7.235 & 9.458 & 6.673 \\
    Test4
    & indoor--parking mix
    & 20:10 & 1.230
    & 12.481 & Fail. & 5.865 & \textbf{2.904} & 5.064 \\
    Test5
    & aggressive rotation
    & 03:00 & 0.116
    & 0.070 & Fail. & \textbf{0.005} & 0.095 & \textbf{0.006} \\
    Test6
    & small indoor room
    & 02:09 & 0.061
    & 0.074 & Fail. & \textbf{0.035} & \textbf{0.033} & \textbf{0.032} \\
    Test7
    & campus e-bike
    & 23:56 & 4.446
    & 79.631 & 153.910 & \textbf{50.547} & 53.690 & 57.166 \\
    \bottomrule
  \end{tabular*}
\end{table*}

\subsection{Experimental Setup}
\label{subsec:exp_setup}

The evaluation uses the LiDAR--IMU instantiation detailed in Section~\ref{sec:lio_instantiation}. The benchmark comprises seven self-collected sequences that include a loop corridor, multi-floor stairs, campus walkways, an indoor--parking route, a small indoor room, an e-bike traversal, and an aggressive-rotation sequence. \texttt{Test1} is the primary diagnostic sequence: a 418~m handheld traversal through a corridor whose repetitive geometry induces weakly observable directions. The remaining six sequences characterize the operating boundary of the same instantiation beyond this diagnostic setting.

End-to-end trajectory error in meters is used as the primary metric in Table~\ref{tab:benchmark_results}. This metric summarizes the terminal discrepancy of each run, but it does not by itself measure local map consistency, latency, or closed-loop behavior. Cases where a method diverges or produces a non-viable estimate are marked as \emph{Fail.}, with no numerical error assigned.

The comparison includes four representative LiDAR--IMU baselines: FAST-LIO2~\cite{xu_fast-lio2_2022}, LIO-SAM~\cite{shan_lio-sam_2020}, Point-LIO~\cite{he_pointlio_2023}, and Faster-LIO~\cite{bai_faster-lio_2022}. All methods process the same raw sensor streams with loop closure and global optimization disabled. No sequence-specific parameter tuning is used unless noted. This protocol supports a controlled comparison on the reported sequences, but it should not be interpreted as a statistical claim that one method is better across all LiDAR--IMU operating domains.

\subsection{Primary Diagnostic: Degenerate Corridor}
\label{subsec:corridor_main}

The 418~m loop corridor is used as the primary diagnostic scenario for directional degeneracy. In this environment, dominant planar surfaces provide stronger constraints transverse to the walls than along the corridor direction. The sequence therefore tests whether the LiDAR-triggered update can be regularized when the local information matrix is ill-conditioned.

The first row of Table~\ref{tab:benchmark_results} reports the corridor result. The FUSE instantiation obtains a 1.626~m end-to-end error, compared with 1.765~m for Faster-LIO, 8.189~m for FAST-LIO2, and 8.077~m for Point-LIO, while LIO-SAM does not return a viable estimate. Relative to Faster-LIO, the strongest non-FUSE baseline on this sequence, this corresponds to a 7.9\% lower end-to-end error. These numbers support the evaluated instantiation under the corridor degeneracy setting; they do not imply categorical dominance across all geometric contexts.

\begin{figure*}[t]
  \centering
  \subfloat[FAST-LIO2: top-down view with local detail.]{
    \includegraphics[width=0.48\textwidth]{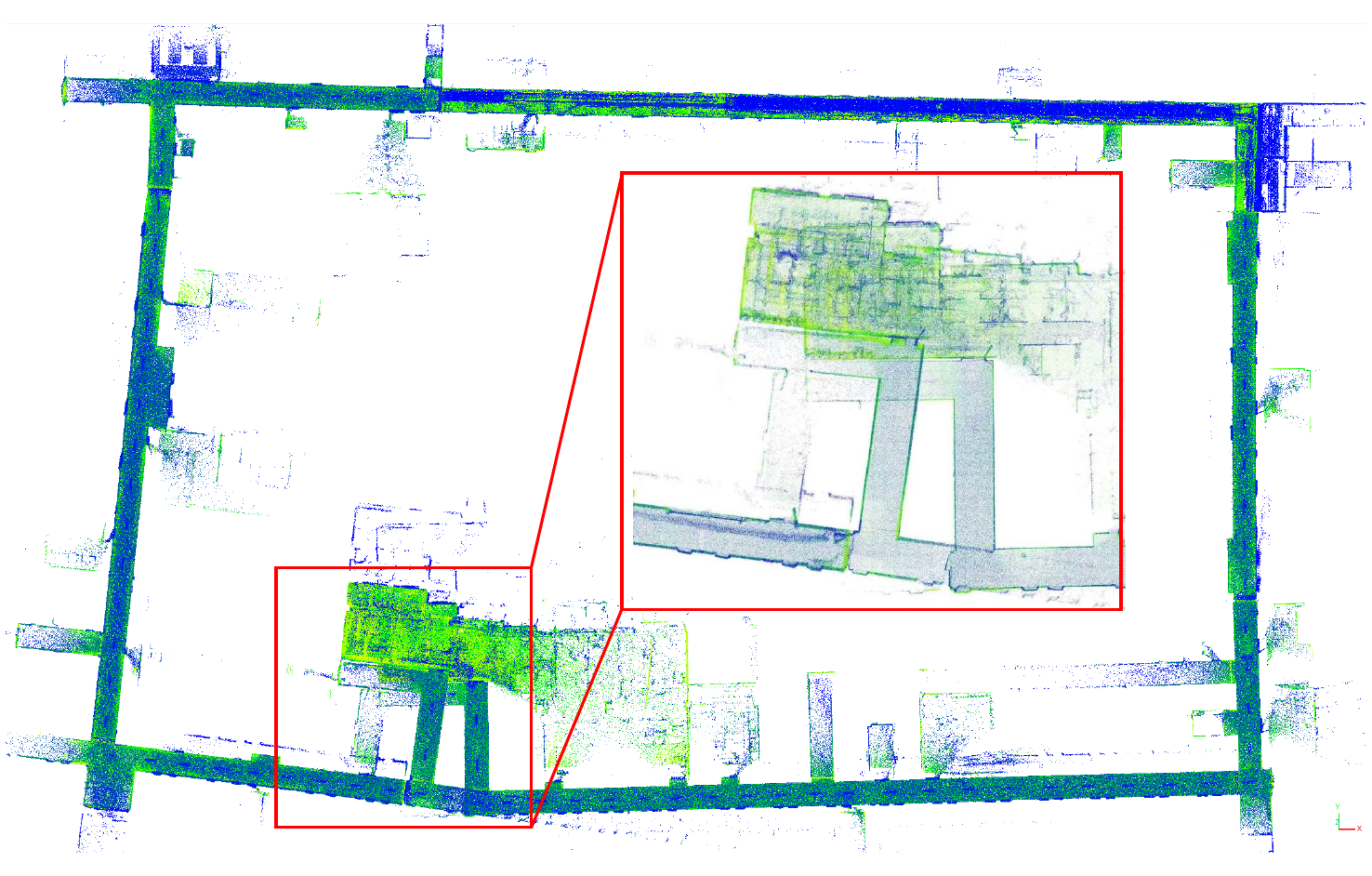}
    \label{fig:corridor_fastlio2_top}
  }\hfill
  \subfloat[FUSE: top-down view with local detail.]{
    \includegraphics[width=0.48\textwidth]{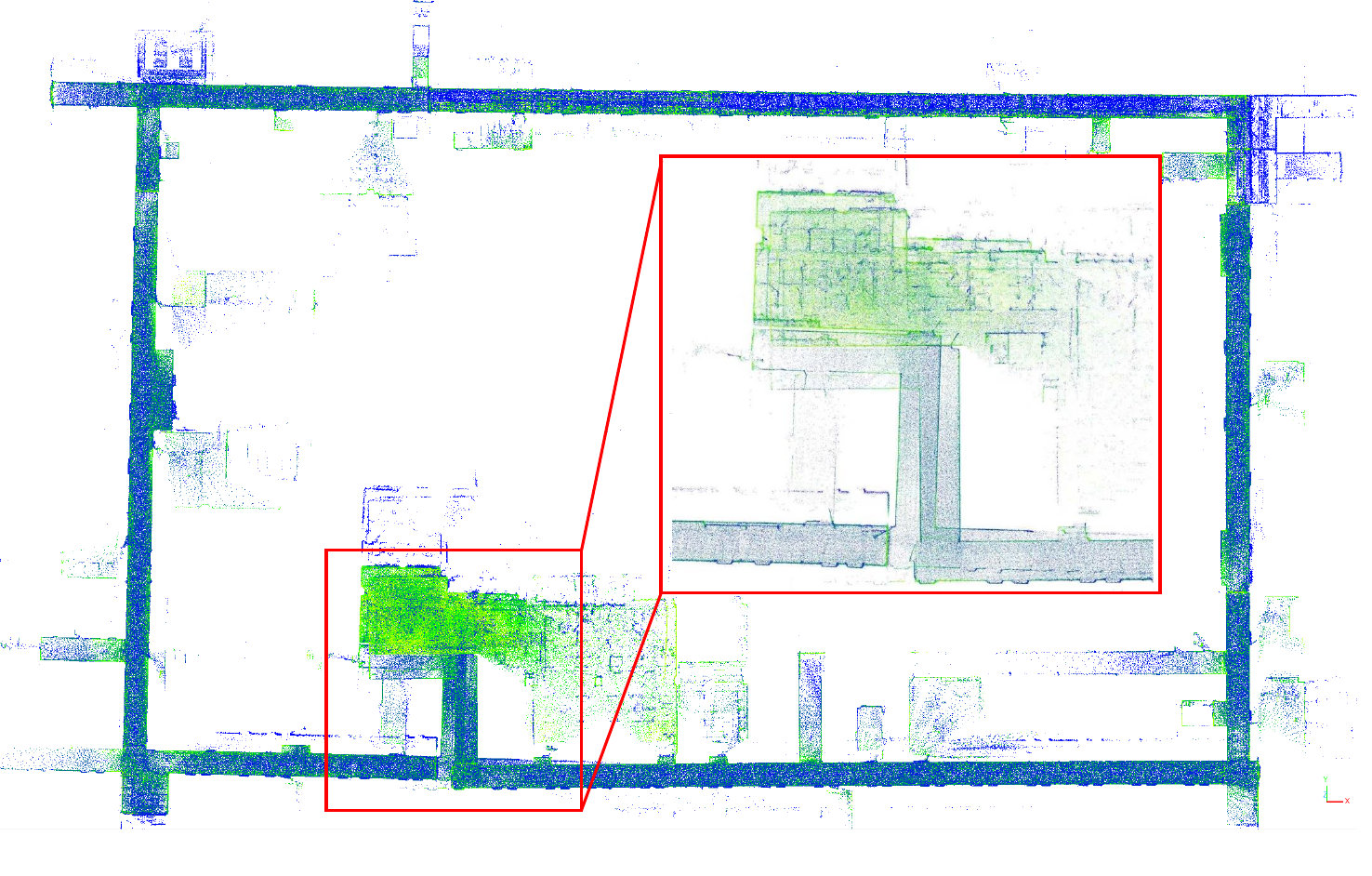}
    \label{fig:corridor_fuse_top}
  }\\[0.35em]
  \subfloat[FAST-LIO2: oblique 3D view.]{
    \includegraphics[width=0.48\textwidth]{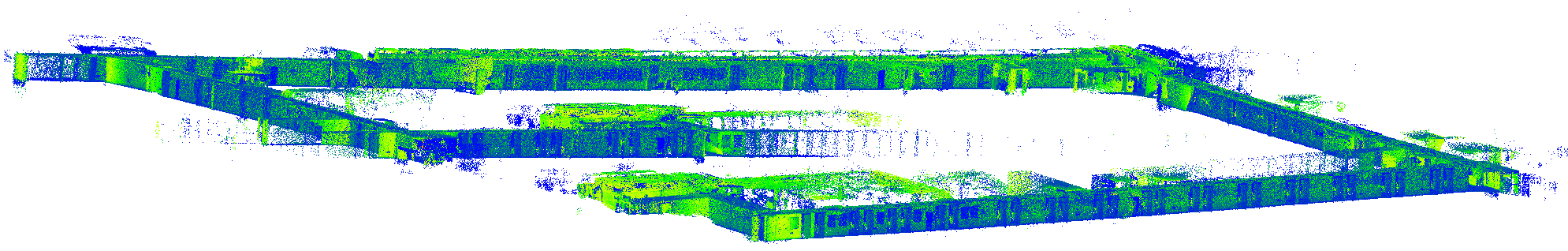}
    \label{fig:corridor_fastlio2_3d}
  }\hfill
  \subfloat[FUSE: oblique 3D view.]{
    \includegraphics[width=0.48\textwidth]{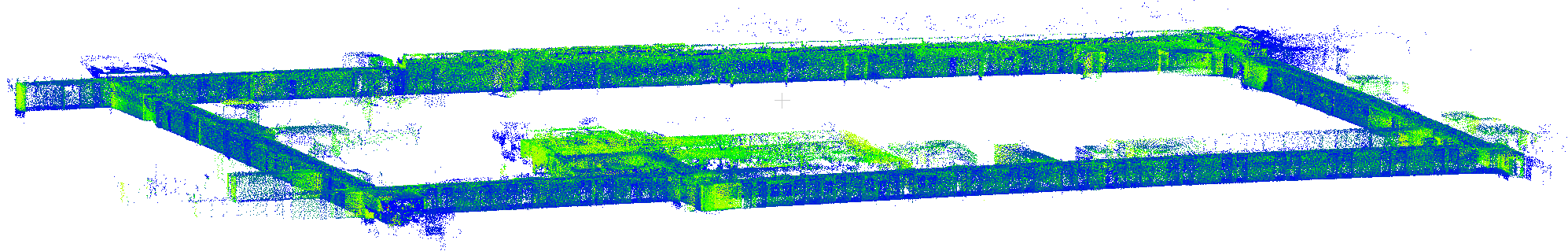}
    \label{fig:corridor_fuse_3d}
  }
  \caption{Primary qualitative comparison on the loop-corridor benchmark (\texttt{Test1}). (a)--(b) Top-down views with local detail; (c)--(d) oblique 3D views. The comparison shows representative map structure and loop behavior for FAST-LIO2 and the FUSE LiDAR--IMU instantiation under strongly directional corridor geometry.}
  \label{fig:corridor_map}
\end{figure*}

Qualitative evidence is presented in Figure~\ref{fig:corridor_map}. The top-down and oblique views illustrate the map structure and loop behavior obtained by FAST-LIO2 and by the FUSE LiDAR--IMU instantiation on the corridor sequence. These visualizations are used as supporting evidence and should be interpreted together with the numerical results in Table~\ref{tab:benchmark_results}.

The corridor result is consistent with the realization-level mechanisms introduced in Section~\ref{sec:lio_instantiation}. IMU propagation maintains queryable temporal states between LiDAR frames; local association and residual screening control which geometric constraints enter the update; and the degeneracy-aware update attenuates corrections along weakly observable tangent-space directions. The ablation study below isolates these mechanisms on \texttt{Test1}; the claim is limited to the evaluated configuration.

\subsection{Broader Operating Evidence}
\label{subsec:benchmark_wide}

The remaining six sequences are used to characterize the operating boundary of the current LiDAR--IMU instantiation. They include vertical multi-floor motion, campus-scale traversal, mixed indoor--parking motion, small-room operation, e-bike motion, and aggressive rotation. The purpose of this broader set is to report where the instantiation performs competitively and where other baselines obtain lower end-to-end errors.

Table~\ref{tab:benchmark_results} shows a non-uniform ranking across these sequences. Under the centimeter-level tie criterion, FUSE belongs to the leading group on the multi-floor stairs, aggressive-rotation, and small indoor-room sequences. In larger or mixed environments, the ranking changes: FUSE reports 6.673~m on the campus walkway, 5.064~m on the indoor--parking route, and 57.166~m on the campus e-bike sequence, while LIO-SAM, Faster-LIO, or Point-LIO report lower errors on these respective sequences. These results indicate competitive behavior in several structured settings and also define the operating boundary of the present instantiation.

\begin{figure*}[t]
  \centering
  \subfloat[multi-floor stairs: 3D trajectory comparison.]{
    \includegraphics[width=0.48\textwidth]{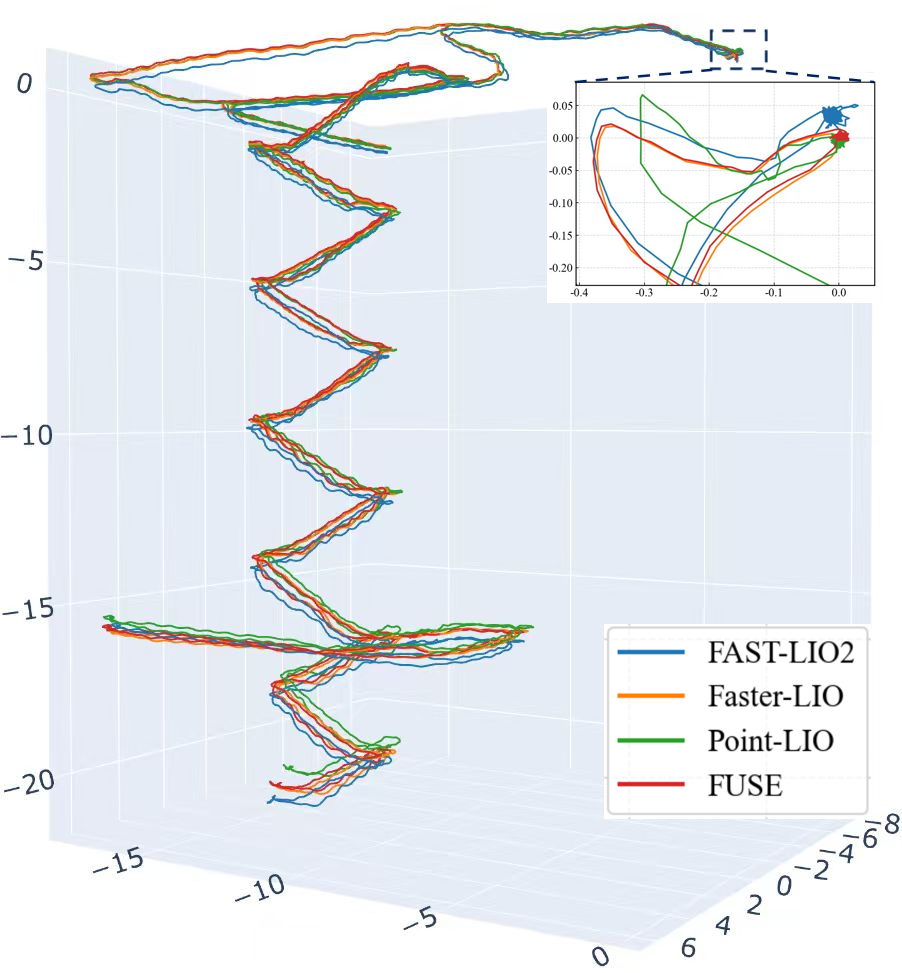}
    \label{fig:stairs_traj_3d}
  }\hfill
  \subfloat[multi-floor stairs: representative local map view.]{
    \includegraphics[width=0.48\textwidth]{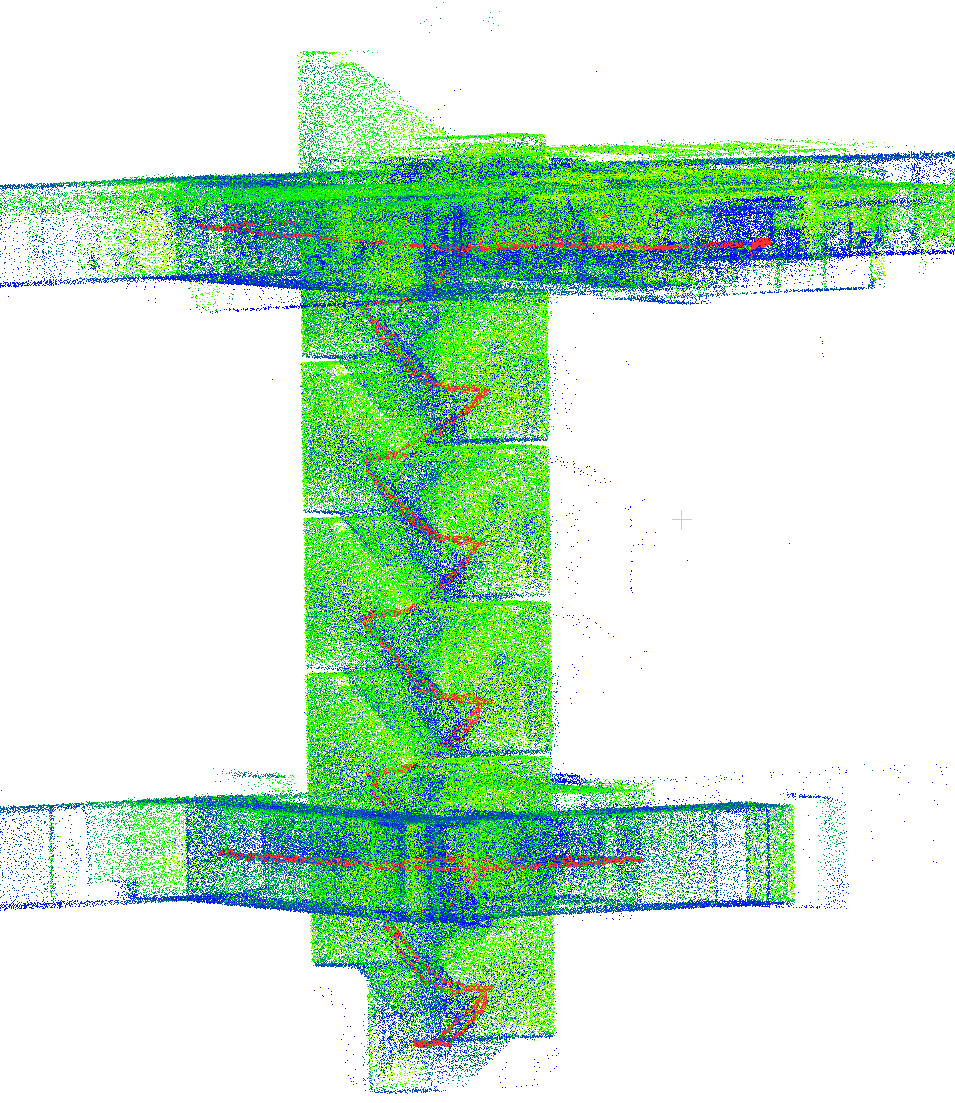}
    \label{fig:stairs_map}
  }\\[0.45em]
  \subfloat[campus e-bike: top-view trajectory comparison.]{
    \includegraphics[width=0.48\textwidth]{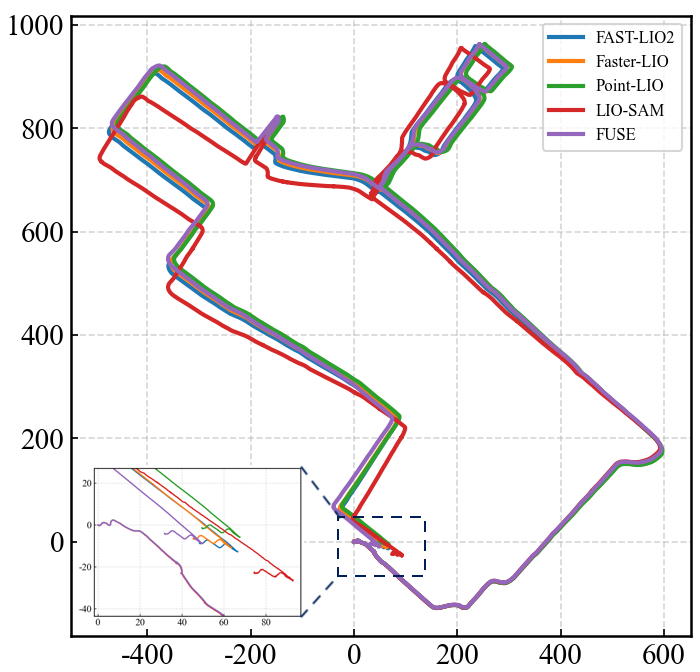}
    \label{fig:campus_traj_top}
  }\hfill
  \subfloat[campus e-bike: representative local map view.]{
    \includegraphics[width=0.48\textwidth]{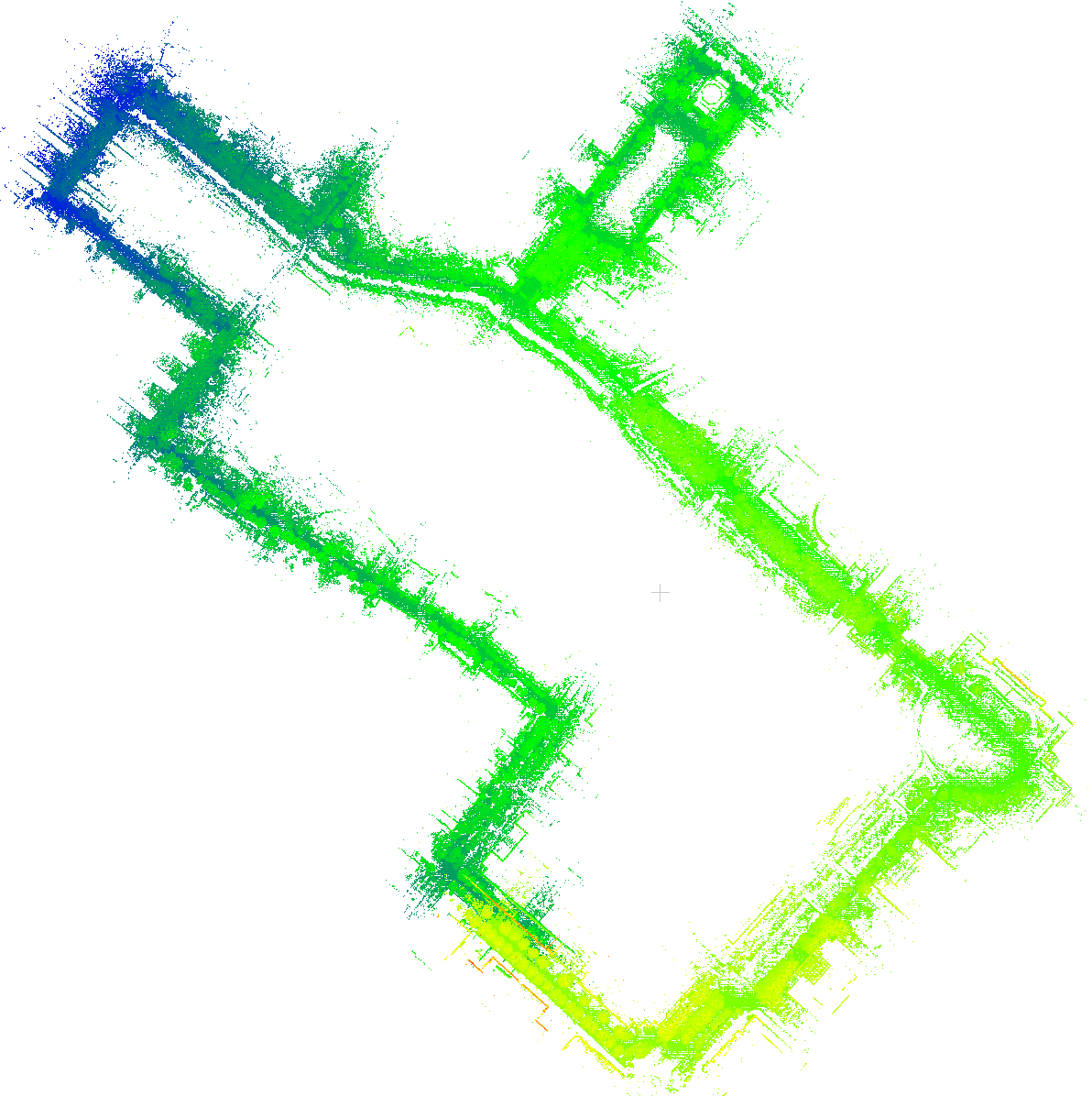}
    \label{fig:campus_map}
  }
  \caption{Qualitative evaluation across additional operating regimes. The multi-floor staircase sequence (top) shows sustained 3D motion and a representative local map; the campus e-bike traversal (bottom) shows a top-view trajectory comparison and representative map output during extended outdoor motion.}
  \label{fig:multi_sequence_results}
\end{figure*}

Qualitative assessments for non-corridor scenarios are presented in Figure~\ref{fig:multi_sequence_results}. The staircase sequence (Fig.~\ref{fig:stairs_traj_3d}--\ref{fig:stairs_map}) shows the trajectory and representative map output during sustained three-dimensional motion and vertical transitions. The campus e-bike traversal (Fig.~\ref{fig:campus_traj_top}--\ref{fig:campus_map}) shows representative output during extended outdoor motion. These visual examples provide context for the quantitative results but do not replace the non-uniform ranking reported in Table~\ref{tab:benchmark_results}.

\begin{figure*}[t]
  \centering
  \subfloat[Second Gate.]{
    \includegraphics[width=0.236\textwidth]{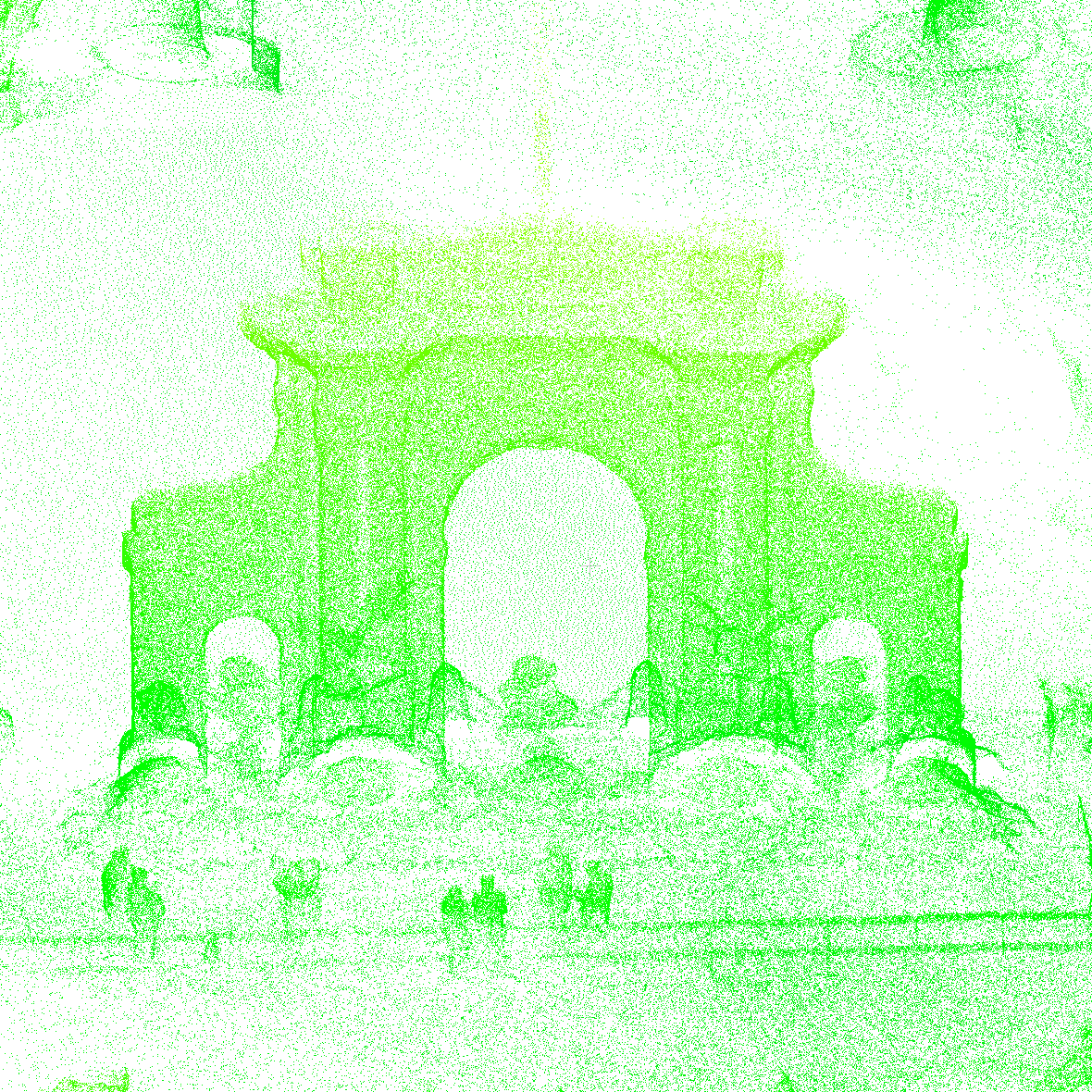}
    \label{fig:campus_second_gate}
  }\hfill
  \subfloat[Tsinghua Xuetang.]{
    \includegraphics[width=0.236\textwidth]{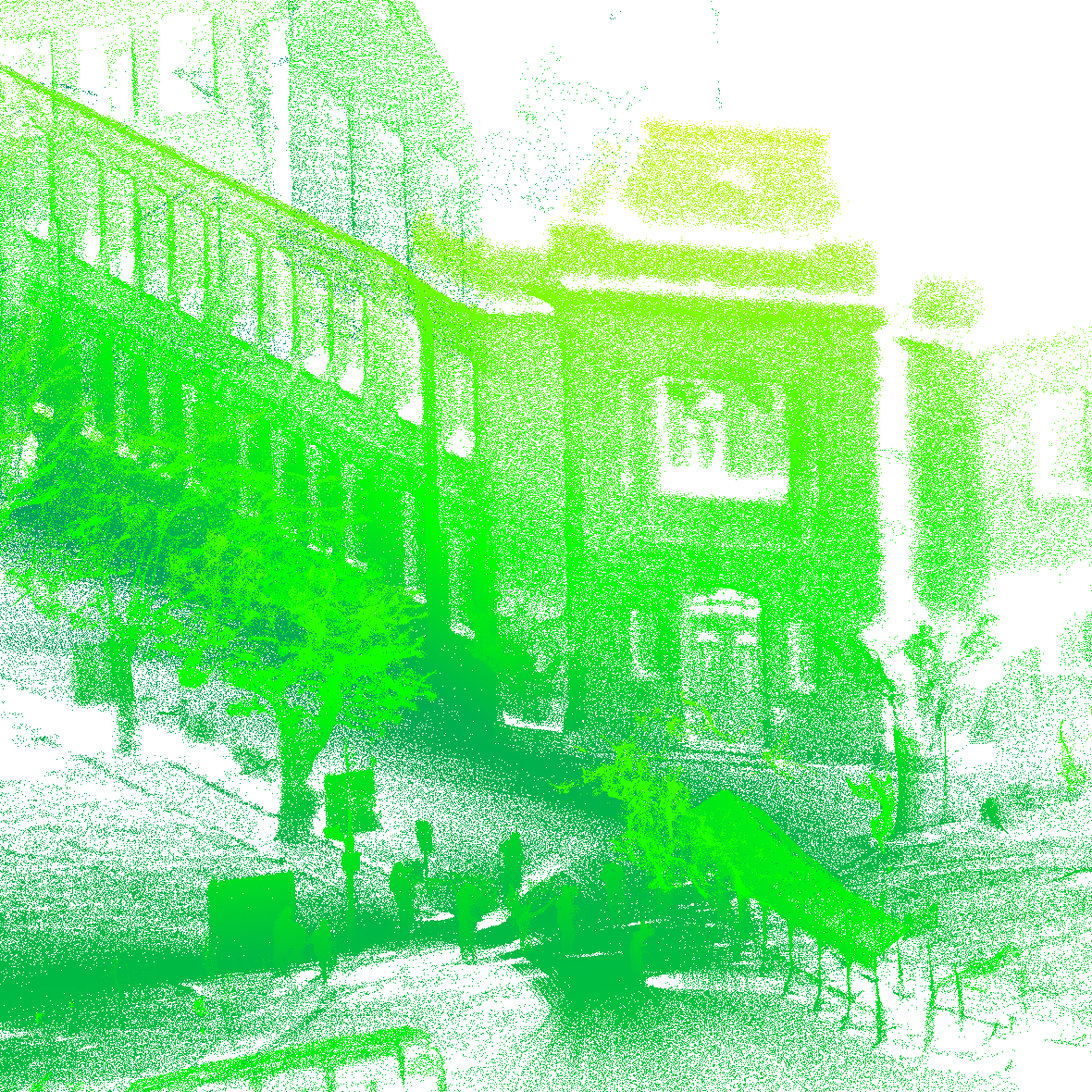}
    \label{fig:campus_xuetang}
  }\hfill
  \subfloat[Grand Auditorium.]{
    \includegraphics[width=0.236\textwidth]{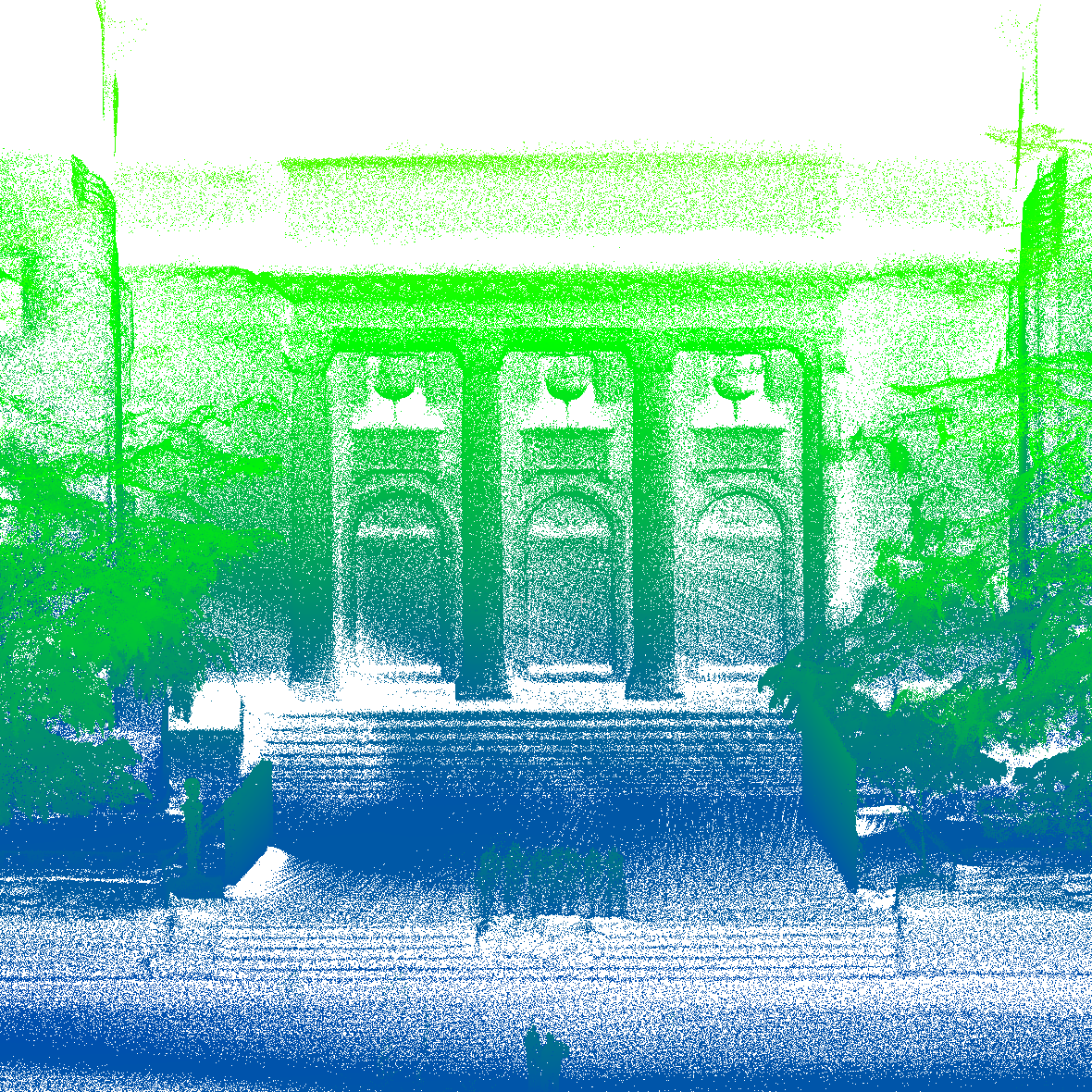}
    \label{fig:campus_auditorium}
  }\hfill
  \subfloat[Lecture Hall.]{
    \includegraphics[width=0.236\textwidth]{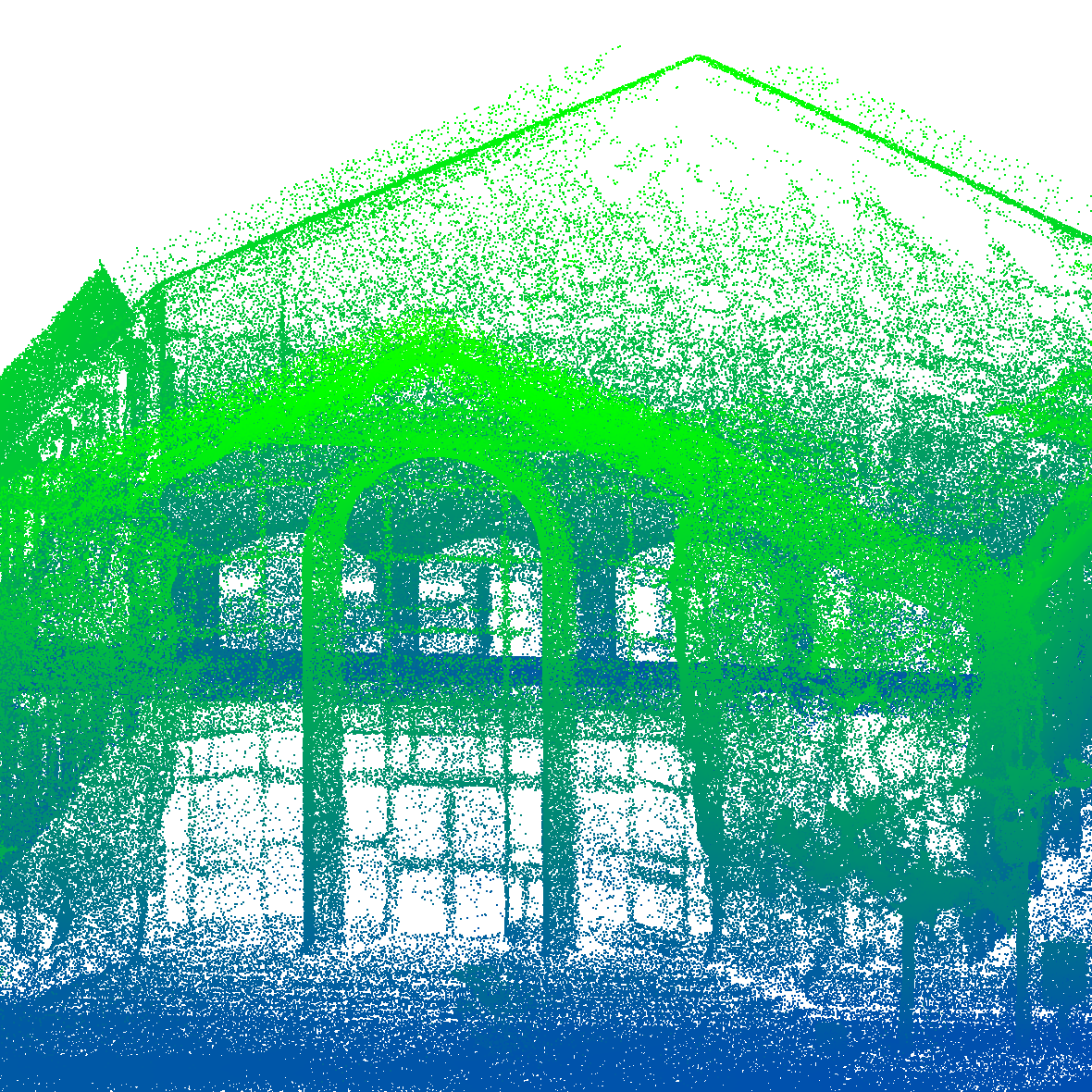}
    \label{fig:campus_lecture_hall}
  }
  \caption{Representative landmark-level reconstructions generated by the FUSE LiDAR--IMU instantiation during the campus traversal. The views illustrate building-scale map structure for several campus landmarks.}
  \label{fig:campus_landmark_maps}
\end{figure*}

Figure~\ref{fig:campus_landmark_maps} provides representative landmark-level reconstructions from the campus traversal. The views of the Second Gate of Tsinghua University, Tsinghua Xuetang, the Grand Auditorium, and the Lecture Hall illustrate the building-scale map structure obtained by the current instantiation.

Collectively, the broader results show that the current instantiation remains viable across the reported sequences, but they also show that the lowest-error method depends on the scene. This non-uniform ranking is important for the scope of the paper: the experiments support the organization and realization-level mechanisms of FUSE in the evaluated LiDAR--IMU setting, not broad dominance over specialized systems.

\subsection{Ablation Studies}
\label{subsec:corridor_ablation}

The ablation study uses the loop-corridor sequence (\texttt{Test1}) because its directional degeneracy directly exercises the mechanisms introduced in Section~\ref{sec:lio_instantiation}. The variants remove degeneracy-aware update gating, residual screening, or the covariance-propagation setting from the current LiDAR--IMU realization. These mechanisms are evaluated as realization-level safeguards, not as additional framework layers.

\begin{table*}[t]
  \centering
  \caption{Mechanistic ablation study on the degenerate loop-corridor sequence (\texttt{Test1}). End-to-end trajectory error and terminal drift components \((\Delta x,\Delta y,\Delta z)\) are reported in meters. The scalar error is the Euclidean norm of the terminal drift vector. Values are reported to three decimals for numerical consistency with Table~\ref{tab:benchmark_results}; the comparison emphasizes differences among variants rather than millimeter-level measurement accuracy.}
  \label{tab:corridor_ablation}
  \small
  \setlength{\tabcolsep}{3pt}
  \renewcommand{\arraystretch}{1.10}
  \begin{tabular*}{\textwidth}{@{\extracolsep{\fill}}
      p{0.27\textwidth}
      >{\centering\arraybackslash}p{0.09\textwidth}
      >{\centering\arraybackslash}p{0.075\textwidth}
      >{\centering\arraybackslash}p{0.09\textwidth}
      >{\centering\arraybackslash}p{0.090\textwidth}
      >{\centering\arraybackslash}p{0.150\textwidth}
      >{\centering\arraybackslash}p{0.080\textwidth}
      >{\centering\arraybackslash}p{0.080\textwidth}
    }
    \toprule
    \makecell[l]{Variant}
    & \makecell[c]{Deg.-aware\\ update}
    & \makecell[c]{Residual\\ screening}
    & \makecell[c]{Covariance\\ propagation}
    & \makecell[c]{End-to-end\\ error (m)}
    & \makecell[c]{Drift components\\ $(\Delta x,\Delta y,\Delta z)$ (m)}
    & \makecell[c]{Dominant\\ drift axis}
    & \makecell[c]{Least-drift\\ axis} \\
    \midrule
    FAST-LIO2
    & -- & -- & Baseline
    & 8.189
    & $(-3.553,\ 2.058,\ 7.085)$
    & $Z$
    & $Y$ \\

    FUSE w/o degeneracy-aware update
    & \ding{55} & \ding{51} & \ding{51}
    & 2.193
    & $(-0.887,\ 1.764,\ 0.956)$
    & $Y$
    & $X$ \\

    FUSE w/o residual screening
    & \ding{51} & \ding{55} & \ding{51}
    & 3.531
    & $(-0.706,\ 2.677,\ 2.192)$
    & $Y$
    & $X$ \\

    FUSE w/o cov-propagation setting
    & \ding{51} & \ding{51} & \ding{55}
    & 2.382
    & $(-0.914,\ 1.723,\ 1.367)$
    & $Y$
    & $X$ \\

    FUSE full
    & \ding{51} & \ding{51} & \ding{51}
    & 1.626
    & $(-0.909,\ 1.171,\ -0.669)$
    & $Y$
    & $Z$ \\
    \bottomrule
  \end{tabular*}
\end{table*}

The quantitative results in Table~\ref{tab:corridor_ablation} show that the full FUSE instantiation reports the lowest error among the tested variants. Removing residual screening increases the end-to-end error from 1.626~m to 3.531~m. This is the largest change among the three ablations and indicates that admissibility control is important for this corridor sequence, where unstable local associations can affect the information matrix and the resulting state increment.

Removing the degeneracy-aware update increases the error to 2.193~m, and reverting the covariance-propagation setting increases it to 2.382~m. These results support the use of subspace-specific update attenuation and propagated uncertainty weighting in the evaluated corridor setting. The drift components further indicate that the variants change not only the total end-to-end error but also the direction in which drift accumulates. This analysis supports the realization-level mechanisms used in Section~\ref{sec:lio_instantiation}; it does not imply that the same ablation ordering will hold for all scenes or estimator backends.

\subsection{Runtime Behavior}
\label{subsec:runtime_profile}

We report the runtime profile of the current LiDAR--IMU instantiation under the mixed-rate propagation and update regime. The purpose is to document the computation observed on the reported hardware and sensor configuration, not to establish a general real-time guarantee for all platforms or closed-loop controllers.

\begin{table}[!t]
  \centering
  \caption{Experimental platform of the current FUSE LiDAR--IMU instantiation.}
  \label{tab:exp_platform}
  \small
  \setlength{\tabcolsep}{4pt}
  \renewcommand{\arraystretch}{1.08}
  \begin{tabular}{p{0.34\linewidth} p{0.56\linewidth}}
    \toprule
    Item & Setting \\
    \midrule
    LiDAR            & Livox Mid-360 \\
    IMU              & Built-in IMU of the Livox Mid-360 \\
    LiDAR rate       & 10~Hz \\
    IMU rate         & 200~Hz \\
    Platform         & Intel Core i9-14900 \\
    Collection mode  & Handheld \\
    Operating system & Ubuntu 24.04 \\
    Implementation environment & ROS 2 Humble / C++ \\
    \bottomrule
  \end{tabular}
\end{table}

\begin{table}[!t]
  \centering
  \caption{Runtime profile of the current FUSE LiDAR--IMU instantiation.}
  \label{tab:runtime_profile}
  \small
  \setlength{\tabcolsep}{4pt}
  \renewcommand{\arraystretch}{1.08}
  \begin{tabular}{
      >{\raggedright\arraybackslash}m{0.65\linewidth}
      >{\raggedright\arraybackslash}m{0.25\linewidth}
    }
    \toprule
    Metric & Value \\
    \midrule
    Propagation rate              & 200~Hz \\
    LiDAR update rate             & 10~Hz \\
    Single-core CPU usage         & $\sim 12\%$ \\
    Mean LiDAR update time        & 11.7~ms \\
    P95 LiDAR update time         & 17.7~ms \\
    Max LiDAR update time         & 30.1~ms \\
    End-to-end latency            &
    \begin{tabular}[c]{@{}l@{}}
      propagation: $<1$~ms\\
      update: 13~ms
    \end{tabular} \\
    Local association structure   & iVox \\
    Corrected-state output rate   & 10~Hz \\
    \bottomrule
  \end{tabular}
\end{table}

The hardware configuration and software environment are summarized in Table~\ref{tab:exp_platform}. Table~\ref{tab:runtime_profile} reports approximately 12\% single-core CPU usage, a mean LiDAR update time of 11.7~ms, a P95 update time of 17.7~ms, and a maximum update time of 30.1~ms. These update times are below the 100~ms interval of the reported 10~Hz LiDAR stream on the Intel Core i9-14900 platform. The propagation path reports sub-millisecond latency, while the corrected-state output rate is 10~Hz.

These measurements characterize the reported implementation and hardware configuration. They indicate that the evaluated instantiation fits within the timing budget of the tested LiDAR stream, but they do not establish behavior across sensors, processors, spatial indexes, or estimator backends.

\subsection{Limitations}

The presented evaluation covers one LiDAR--IMU instantiation of FUSE and therefore samples only part of the framework design space. The current study does not exhaustively compare estimator formulations such as KF, ESKF, IESKF, and NANO-style updates, nor does it evaluate all sensor combinations, spatial indexing strategies, or map-update policies that could be implemented behind the same state-estimation interface. The analysis is focused on the realization-level mechanisms of the reported LiDAR--IMU implementation rather than on exhaustive validation of all framework interfaces.

The 418~m loop corridor is a useful diagnostic for directional degeneracy, but it is not a comprehensive proxy for all degenerate or large-scale environments. The non-uniform ranking across the broader sequences in Table~\ref{tab:benchmark_results} shows that the current instantiation is not the lowest-error method in every setting. The runtime profile is also context-dependent because it reflects the Intel Core i9-14900 platform, the Livox Mid-360 sensor, the reported implementation, and the selected local association structure.

Future work should therefore evaluate additional estimator formulations under the same state-estimation interface, compare spatial indexes for local geometric association, and study localization-only map-update policies. The vehicular and robotic scope of the framework should be read as a state-estimation organization claim rather than as exhaustive validation across all driving and robotic operating regimes. Closed-loop deployment on vehicular and robotic platforms is also needed to assess how the interface and runtime behavior affect time-sensitive control tasks.

\section{Conclusion}

This paper presented FUSE, a framework for unified state estimation in vehicular and robotic SLAM systems. The framework addresses a common source of rigidity in tightly coupled SLAM: temporal processing, local geometric association, estimator formulation, and map-update policy are often specified together inside a method-specific pipeline. FUSE exposes observation ingestion, propagation, update, and query through a state-estimation interface, so that temporal scheduling, residual-ready local association, posterior-update formulation, and map persistence can be organized as separate interface roles.

A LiDAR--IMU instantiation was used to evaluate this organization under mixed-rate sensing and directional degeneracy. On the 418~m loop-corridor sequence, the instantiation reports a 1.626~m end-to-end trajectory error, corresponding to a 7.9\% reduction relative to Faster-LIO, the lowest-error non-FUSE baseline on that sequence. The ablation study further shows that residual screening, degeneracy-aware update, and covariance propagation affect the reported corridor result. The broader benchmark and runtime measurements define the current operating boundary rather than proving broad dominance. Future work will extend this evaluation to additional estimator backends, spatial indexes, sensor combinations, map-update policies, and closed-loop deployments on vehicular and robotic platforms.

\section*{Acknowledgment}

The authors gratefully acknowledge Jianqiu Wang and Gang Huang from China Mobile (Shanghai) Industrial Research Institute; Hongbo Li from Geekplus Technology Co., Ltd.; Mingming Wang and Weiqiang Liang from Guangzhou Automobile Group Co., Ltd.; Honglin Li and Tao Chen from Dongfeng Motor Corporation Research and Development Institute; Jiuhua Zhao and Hui Guo from IM Motors Technology Co., Ltd.; Ji Tao and Gang He from Chongqing Changan Automobile Co., Ltd.; Qian Zhang from Horizon Robotics; Kehua Sheng and Bo Zhang from DiDi Voyager Labs, DiDi Autonomous Driving; Beiyan Jiang and Tianyi Zhang from the School of Vehicle and Mobility, Tsinghua University; and Haoning Wang from the Department of Mathematical Sciences, Tsinghua University, for their valuable discussions, technical support, and assistance with experimental verification.

{
  \balance
  \bibliographystyle{IEEEtran}
  \bibliography{papers}
}

\end{document}